\documentclass[acmsmall, screen, nonacm]{acmart}
\AtBeginDocument{%
  }

\setcopyright{acmlicensed}
\copyrightyear{2025}
\acmYear{2025}
\acmDOI{XXXXXXX.XXXXXXX}

\acmJournal{TOSEM}
\acmVolume{1}
\acmNumber{1}
\acmArticle{1}
\acmMonth{1}

\usepackage{array}
\usepackage{makecell} 
\usepackage{multirow}
  
\newcolumntype{L}[1]{>{\raggedright\arraybackslash}m{#1}} 
\newcolumntype{S}[1]{>{\scriptsize\raggedright\arraybackslash}m{#1}}

\usepackage[most]{tcolorbox}
\usepackage{lipsum}
\newtcolorbox{conclusion}[1]{%
    title={Conclusion - #1}, 
    colback=green!5!white,
    colframe=green!60!black,
    colbacktitle=green!60!black,
    coltitle=white,
    fonttitle=\bfseries,
    breakable 
}

\begin{document}

\title{Robust, Observable, and Evolvable Agentic Systems Engineering: A Principled Framework Validated via the Fairy GUI Agent}

\author{Jiazheng Sun}
\email{jzsun24@m.fudan.edu.cn}
\orcid{0009-0009-2107-1100}
\affiliation{%
  \institution{Fudan University}
  \city{Shanghai}
  \country{China}
}

\author{Ruimeng Yang}
\orcid{0009-0002-2778-2121}
\affiliation{%
  \institution{Fudan University}
  \city{Shanghai}
  \country{China}
}

\author{Xu Han}
\orcid{0009-0005-2664-5637}
\affiliation{%
  \institution{Fudan University}
  \city{Shanghai}
  \country{China}
}

\author{JiaYang Niu}
\orcid{0009-0004-5778-3912}
\affiliation{%
  \institution{Fudan University}
  \city{Shanghai}
  \country{China}
}

\author{Mingxuan Li}
\affiliation{%
  \institution{Fudan University}
  \city{Shanghai}
  \country{China}
}

\author{Te Yang}
\affiliation{%
  \institution{Fudan University}
  \city{Shanghai}
  \country{China}
}

\author{Yongyong Lu}
\orcid{0009-0006-5346-5225}
\affiliation{%
  \institution{Fudan University}
  \city{Shanghai}
  \country{China}
}

\author{Xin Peng}
\authornote{Corresponding author}
\orcid{0000-0003-3376-2581}
\email{xinpeng@fudan.edu.cn}
\affiliation{%
  \institution{Fudan University}
  \city{Shanghai}
  \country{China}
}

\renewcommand{\shortauthors}{Sun et al.}

\begin{abstract}
The Agentic Paradigm faces a significant Software Engineering Absence, yielding Agentic systems commonly lacking robustness, observability, and evolvability. To address these deficiencies, we propose a principled engineering framework comprising Runtime Goal Refinement (RGR), Observable Cognitive Architecture (OCA), and Evolutionary Memory Architecture (EMA). In this framework, RGR ensures robustness and intent alignment via knowledge-constrained refinement and human-in-the-loop clarification; OCA builds an observable and maintainable white-box architecture using component decoupling, logic layering, and state-control separation; and EMA employs an execution-evolution dual-loop for evolvability. We implemented and empirically validated Fairy, a mobile GUI agent based on this framework. On RealMobile-Eval, our novel benchmark for ambiguous and complex tasks, Fairy outperformed the best SoTA baseline in user requirement completion by 33.7\%. Subsequent controlled experiments, human-subject studies, and ablation studies further confirmed that the RGR enhances refinement accuracy and prevents intent deviation; the OCA improves maintainability; and the EMA is crucial for long-term performance. This research provides empirically validated specifications and a practical blueprint for building reliable, observable, and evolvable Agentic AI systems.
\end{abstract}



\received{20 February 2007}
\received[revised]{12 March 2009}
\received[accepted]{5 June 2009}

\maketitle

\section{Introduction}
Driven by large language models (LLMs), the field of software engineering is undergoing a profound paradigm shift. Traditional software systems, which function as passive tools, are giving way to a novel Agentic Paradigm. In this new paradigm, systems are expected to exhibit unprecedented levels of automation and adaptability, evolving into goal-oriented agents capable of autonomous planning, reasoning, and executing complex tasks. This evolution enables the high-level automation of complex business processes that were previously human-reliant. However, this fundamental transition—from deterministic, instruction-following programs to non-deterministic, goal-oriented agents—is posing fundamental challenges to core engineering properties such as system robustness and observability. Meanwhile, traditional software engineering theories and practices are proving inadequate to address this shift.

The Agentic field is currently facing a severe challenge: the lack of a systematic software engineering framework. Although researchers have explored high-level behavioral paradigms, such as ReAct \cite{Yao2023ReAct} and Reflexion \cite{Shinn2023Reflexion}, and development toolkits like LangChain \cite{Chase2022LangChain}, AutoGen \cite{Wu2023AutoGen}, Voyager \cite{wang2023voyager},ChatDev \cite{qian2024chatdev},and MemOS \cite{Li2025MemOS}, the evolution of these solutions has been fragmented. The former are behavioral guides, lacking engineering rigor; the latter provide implementation support but do not constitute a normative methodology. This absence of engineering specification has forced development practices to rely heavily on ad-hoc methods. Developers, in turn, are forced to reinvent core SE concepts in each project, leading to a fragmented landscape of heterogeneous, functionally-coupled, and non-observable black-box implementations \cite{doshi2017towards}. Thus, systems built this way, when confronted with real-world complexity, are typically fragile, opaque, and difficult to manage. We argue that these issues stem not from the capability limitations of the underlying LLMs, but fundamentally from an \textbf{Absence of Software Engineering} during system construction. This absence undermines the reliability and trustworthiness essential for agent deployment, becoming a critical barrier to their large-scale application.

This ad-hoc, fragmented development approach manifests in three core defects: \textbf{poor Robustness, insufficient Observability, and a lack of Evolvability}. First, in terms of Robustness, even SoTA agents exhibit significant fragility when confronted with the ambiguity of real-world user instructions and the complexity of multi-stage tasks. Lacking explicit constraints, they tend to perform Blind Refinement—speculatively guessing user intent, causing execution trajectories to deviate frequently from the intended goals and undermining system reliability and user trust. Second, in terms of Observability, the internal workings of these systems are typically a black-box, as their architectures are often characterized by tightly-coupled cognitive components and opaque communication. This architectural opacity, compounded by the inherent non-determinism of LLMs, makes these systems extremely difficult to debug, maintain, and extend. Finally, in terms of Evolvability, existing agents often lack mechanisms for systematically learning from experience. Without a formal process to consolidate successful task experiences into long-term memory, they are relegated to being eternal novices, unable to continually optimize their strategies, which is a fatal flaw for systems required to operate autonomously in dynamic environments.

To address this critical challenge, this paper proposes a comprehensive engineering framework designed to systematically tackle the three aforementioned core defects. The framework consists of three complementary methodologies: \textbf{1) Runtime Goal Refinement (RGR)}, which aims to ensure system Robustness by injecting the rigor of requirements engineering into the runtime and establishing structured human-in-the-loop clarification mechanisms. \textbf{2) Observable Cognitive Architecture (OCA)}, which aims to ensure Observability and maintainability by constructing white-box systems through architectural specifications, such as component decoupling, state-control separation, and logical layering. \textbf{3) Evolvable Memory Architecture (EMA)}, which aims to ensure system Evolvability by formalizing the experiential learning process via an execute-evolve dual-loop model that transforms task experience into reusable knowledge.

We conducted a rigorous empirical study comparing Fairy against multiple SoTA baselines in the Mobile GUI Agent domain to systematically evaluate the performance benefits derived from adhering to the RGR, OCA, and EMA principles. We first assessed Fairy's foundational capabilities on the public AndroidWorld \cite{Rawles2024AndroidWorld} benchmark, followed by an in-depth evaluation on RealMobile-Eval, a benchmark we specifically designed to measure agent performance in handling tasks characterized by high-level complexity and user intent ambiguity. The results provide strong evidence for our framework's effectiveness: Fairy achieved a 33.7\% higher user requirement completion rate compared to the best-performing baseline. Furthermore, a study with human participants demonstrated that architectures adhering to OCA principles significantly enhance system maintainability, substantially reducing the time required for expert developers to extend the system. These controlled comparisons and ablation studies clearly establish a strong causal link between the engineering principles advocated by our framework and the observed improvements in system robustness, observability, and evolvability.

The main contributions of this paper are as follows:

\begin{itemize}
  \item \textbf{A novel and empirically validated engineering framework.} We propose a systematic SE framework—comprising RGR, OCA, and EMA—to address the software engineering absence in Agentic AI development. This framework facilitates the construction of robust, observable, and evolvable agent systems. Empirical studies have demonstrated its effectiveness in enhancing both task performance and maintainability.
  
  \item \textbf{A complete application instance of the framework.} We contribute the design and implementation of the Fairy agent. This serves as a comprehensive demonstration of instantiating the theoretical specifications of RGR, OCA, and EMA into a practical engineering implementation. It can be used as a blueprint for applying the framework in real-world domains.
  
  \item \textbf{A SoTA Mobile GUI Agent and evaluation benchmark.} We contribute Fairy, an agent that surpasses SoTA baselines. To overcome the limitations of existing benchmarks in evaluating complex tasks and ambiguous instructions, we constructed the RealMobile-Eval benchmark. Furthermore, we designed an accompanying LLM-based, automated, fine-grained evaluation pipeline.
\end{itemize}

The remainder of this paper is organized as follows. \textbf{Section 2} introduces the background, addressing the concepts of the Prompt Software Crisis and the Software Engineering Absence. \textbf{Section 3} elaborates on our Agentic Engineering Framework, which comprises RGR, OCA, and EMA. \textbf{Section 4} presents the Fairy case study, demonstrating the pathway to engineering this framework. \textbf{Sections 5 and 6} describe our empirical study, covering the experimental design, results analysis, and a discussion of threats to validity. Finally, \textbf{Section 7} reviews related work, and \textbf{Section 8} concludes the paper by summarizing limitations and outlining future work.

\section{Background}

\subsection{The Promptware Crisis}

The emergence of LLMs is profoundly reshaping how software systems are built and interacted with, giving rise to a new system paradigm: Agentic AI. Unlike traditional software, Agentic AI systems are not passive tool programs. Instead, they are Runtime Entities capable of autonomously perceiving, planning, acting, and continuously learning in open environments. However, this leap in capability is not matched by corresponding engineering specifications and systematic development methodologies, which are severely lacking.

Currently, the development of most Agentic AI systems remains in an ad-hoc, trial-and-error-based mode. Lacking systematic software engineering constraints, this approach causes agents to exhibit unpredictable behavior, unclear responsibility boundaries, and high functional coupling when performing complex tasks. Consequently, although these systems possess powerful decision-making capabilities, their performance is unstable. They are difficult to observe, debug, and maintain, failing to meet the requirements for production-grade applications.

This phenomenon is termed the \textbf{Promptware Crisis: Agentic systems rely on fragile natural language prompts, with their logic and behavior concealed within the black-box of semantic generation.} Analogous to the runaway code complexity of the 20th-century Software Crisis, the current challenge is runaway semantic complexity \cite{zhou2023webarena}.Engineers lack the necessary engineering methods to predict, design, and constrain system behavior. Agentic systems have not matured in step with the growth of LLM intelligence. Instead, due to this persistent lack of software engineering, they continually face crises of \textbf{Non-Determinism Loss of Control}, \textbf{Insufficient Observability}, and \textbf{Lack of Adaptability} .

\subsubsection{\textbf{Non-Determinism Loss of Control}}

Non-determinism is an inherent characteristic of LLMs. In an environment lacking systematic engineering constraints, this intrinsic non-determinism is amplified, leading to uncontrollable and unreliable system behavior. This loss of control manifests in two dimensions: First, \textbf{during macro-planning}, when not constrained by explicit knowledge, an agent relies on its pre-trained general knowledge for reasoning, which can easily lead to planning path divergence. Although agents may maintain accuracy on low-abstraction sub-goals, their reliability plummets when facing complex, long-term tasks that require long reasoning chains and multi-component coordination. Second, \textbf{during micro-resolution}, when facing missing information, the agent tends to engage in intent speculation. While this unilateral decision-making maintains execution flow, it almost inevitably causes the execution trajectory to deviate from the user's true intent, thereby undermining system trustworthiness \cite{liu2023agentbench}.

\subsubsection{\textbf{Insufficient Observability}}

Observability is a prerequisite for the reliability, debuggability, and maintainability of modern complex software systems, yet it is generally lacking in current Agentic AI. \textbf{In single-agent systems}, core cognitive functions (e.g., planning, decision-making, reflection) are often highly coupled within a single component (such as a single prompt), rendering it an atomic black-box with invisible internal logic. \textbf{In multi-agent systems}, the problem manifests as opaque coordination. The data flow (state and context) and control flow (component activation) are highly intertwined and tightly coupled. This black-box architecture, when compounded by LLM non-determinism, makes system states and decision paths difficult to trace or audit. This fundamentally undermines debuggability and presents a critical barrier to deploying these systems in production.

\subsubsection{\textbf{Lack of Adaptability}}

Agentic AI systems are expected to learn continuously, yet current ad-hoc implementations generally lack mechanisms for experience accumulation and evolution. This lack of adaptability is twofold: First, due to a lack of engineering constraints, the system's record of its own Cognitive Stack (i.e., the trajectory of its decisions, actions, and observations) is often ephemeral and unstructured. This prevents valuable runtime experience from being effectively preserved for post-task reflection. Second, the system also lacks a formal consolidation process to abstract and solidify these raw experiences into reusable long-term knowledge. This dual structural and procedural deficiency condemns the agent to being an eternal novice, forced to repeat similar errors on the same task. This undermines the system's potential to improve performance and reduce marginal costs through self-evolution.

\subsection{The Failure of Traditional Software Engineering Theories}

In response to the Promptware Crisis, academia and industry have naturally attempted to apply constraints using traditional SE theories \cite{delemos2013selfadaptive2}. \textbf{However, the foundation of these theories—namely, that a system's goals, structure, and feedback mechanisms can be completely modeled and remain stable at Design-Time—no longer holds true under the Agentic AI paradigm.}

To solve the Non-Determinism Loss of Control crisis, researchers have tried introducing GORE \cite{vanLamsweerde2001GORE} (e.g., KAOS \cite{Dardenne1993KAOS}), HTN \cite{Erol1994HTN}, and their runtime extension, Tropos4AS \cite{Bresciani2004Tropos}. However, these theories are rooted in design-time specification; their core relies on a complete goal model pre-defined by human analysts. Whether used for design-time implementation (KAOS) or runtime interpretation (Tropos4AS), this pre-defined paradigm is in direct opposition to the Agentic AI paradigm, which relies on LLMs to dynamically generate goals at runtime. This reliance on pre-defined models renders them incapable of constraining planning divergence or intent speculation, ultimately causing them to fail in resolving this crisis.

To solve the Insufficient Observability crisis, researchers have attempted to adapt architectures from MAS \cite{Wooldridge1995Agents}, BDI \cite{Rao1995BDI}, or MS-HTN. However, these frameworks, born from the symbolic AI era, were designed for predictable agents whose logic is specified. They cannot manage or constrain LLMs, whose decision logic is emergent. Force-fitting these models stifles agent autonomy. Consequently, these traditional architectures cannot provide effective decoupling solutions for cognitively coupled black-boxes or data-flow/control-flow coupling, proving inadequate for the task.

To solve the Lack of Adaptability crisis, the MAPE-K control loop \cite{Kephart2003Autonomic} appears to be an ideal solution. However, the core of traditional MAPE-K is self-adaptation, not self-evolution. Its Knowledge Base (K) is static and used only for deterministic tuning. Even attempts to dynamize the "K" base into a runtime model via the Models@Runtime (M@R) paradigm \cite{blair2009modelsruntime}\cite{Assmann2012ModelsAtRuntime}only transform it from a static model to a dynamic structured model. Neither the traditional static "K" base nor the M@R dynamic model includes a true learning process capable of handling the ephemeral and unstructured Cognitive Stack generated by LLMs. This fundamentally static nature of the "K" base prevents it from solidifying experience, rendering it ineffective at solving the Lack of Adaptability crisis.

\section{SE Framework for Agentic AI Systems}
\label{ch:engineering_framework}

In this chapter, we propose a systematic software engineering framework that provides a set of structured practices for engineering Agentic AI systems. This framework comprises three methodologies: \textbf{Runtime Goal Refinement (RGR)}, \textbf{Observable Cognitive Architecture (OCA)}, and \textbf{Evolutionary Memory Architecture (EMA)}. Together, they form a complete, closed loop covering the entire lifecycle of agentic systems: from \textbf{Design and Build} to \textbf{Execute and Interact}, and finally to \textbf{Reflect and Evolve}. These methodologies integrate cutting-edge work from the AI domain while also drawing upon mature, time-tested disciplines within SE, such as requirements engineering, system architecture, and adaptive systems. Furthermore, they are specifically adapted and deepened to address the unique characteristics of AI-driven systems, notably their non-determinism and inherent adaptivity.

\begin{figure}[t]
  \centering
  \includegraphics[width=\linewidth]{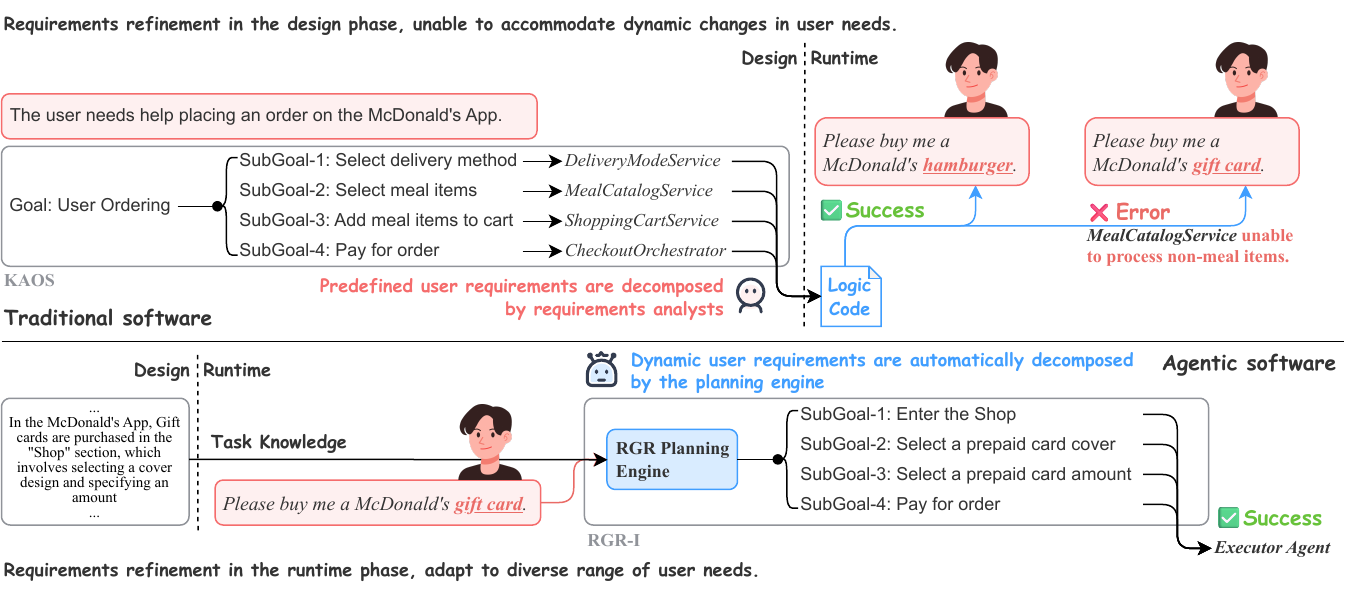}
  \caption{RGR-I (Runtime Refinement) vs. Traditional KAOS (Design-Time Refinement). Traditional software (Top) relies on requirements analysts to map user requirements to system modules at design-time, which renders the system unable to dynamically adapt to user needs; for example, a system targeted at purchasing meal items (e.g., via a MealCatalogService) will fail when receiving a request for a gift card. RGR-I right-shifts requirements refinement for Agentic software (Bottom) to runtime , requiring the planning engine to perform runtime refinement based on Task Knowledge. This allows it to generate new, correct sub-goals and adapt to diverse user needs.}
  \label{fig: rgr1}
\end{figure}

\subsection{The Runtime Goal Refinement (RGR) Methodology}
With the leap in LLM capabilities, software architecture is undergoing a profound evolution toward the Agentic Paradigm. Systems are increasingly expected to transition from passive tool executors to autonomous goal pursuers. This new modality, centered on runtime dynamic planning and autonomous decision-making, renders traditional SE methodologies—which depend on static, design-time specifications—inadequate for addressing requirement ambiguity and environmental volatility \cite{deLemos2013selfadaptive1}. To achieve this runtime flexibility, agent reasoning paradigms such as ReAct were introduced. These paradigms use a Reason-Act-Observe loop to empower agents with dynamic execution capabilities. However, in their pursuit of autonomy, these approaches often sacrifice necessary engineering rigor. In the absence of explicit requirement constraints, agents confronting uncertainty (e.g., decision branches or missing information) are prone to goal deviation and intent speculation, resulting in insufficient behavioral reliability and controllability.

We aim to explore a path that fuses the advantages of both approaches. To this end, we propose the Runtime Goal Refinement (RGR) methodology. RGR is designed to \textbf{right-shift the rigor of requirements engineering from design-time into runtime} (see Figure \ref{fig: rgr1}). This approach seeks to ensure that an agent's execution trajectory remains aligned with the user's true intent, without compromising its flexibility. RGR borrows specification principles from Goal-Oriented Requirements Engineering (GORE) \cite{Kephart2003Autonomic}. It mandates that the agent's process of refining user needs (i.e., planning tasks based on user instructions) be explicitly constrained by both \textbf{Task Knowledge} (a rulebase governing goal decomposition) and \textbf{Environmental Knowledge} (an understanding of available agents and their responsibilities). Furthermore, RGR requires the system to dynamically classify runtime sub-goals into two distinct types: \textbf{Runtime Requirements}, which are clearly specified and can be autonomously executed, and \textbf{Runtime Expectations}, which are underspecified and necessitate user intervention. Upon identifying an Expectation, the system pauses autonomous execution and activates user interaction via an Intent Scaffolding to obtain clarification. Autonomous execution resumes only after this Expectation is refined into an executable Requirement. This establishes a closed-loop, collaborative control flow that balances flexibility and rigor. The RGR is defined by specifications organized into the following two core pillars:

\subsubsection{\textbf{Pillar I: Knowledge-Constrained Goal Refinement and Responsibility Assignment}}

In GORE, goal refinement and responsibility assignment are manual, human-driven cognitive processes. In contrast, within paradigms like ReAct, planning relies on the outcomes of previous actions and observations. RGR fuses concepts from KAOS (a seminal work in GORE) and ReAct. We define \textbf{Goal Refinement} and \textbf{Responsibility Assignment} as a process executed automatically at runtime by a \textbf{Planning Engine}, which is explicitly constrained by both \textbf{Task Knowledge} and \textbf{Environmental Knowledge}.
The example in Figure \ref{fig: rgr1} illustrates the RGR-I goal refinement process, demonstrating how it is constrained by Task Knowledge.

\paragraph{\textbf{RGR Specifications:}}

\subparagraph{\textbf{[RGR-I.1: Implement Planning Engine]:}}
The system must implement one or more Planning Engine components. Their responsibility is to receive a high-level goal and dynamically refine it into a set of sub-goals.

\begin{itemize}
  \item \textbf{Single Component:}
        A single component performs full-stack, recursive refinement, from the highest level (user intent) down to the lowest level (atomic operations).
  \item \textbf{Multi-Component (Hierarchical):}
        The RGR refinement process is distributed. Each component handles refinement at its specific abstraction level and passes sub-goals to the next layer via Cognitive Delegation.
\end{itemize}

\subparagraph{\textbf{[RGR-I.2: Constrain Refinement using Task Knowledge]:}}
Task Knowledge defines the rules, domain knowledge, or method library the engine uses to perform logical decomposition. It must support (or be equivalent to) the core Goal Refinement operations of KAOS to ensure logical completeness. The Planning Engine's refinement behavior cannot be arbitrary; it must be explicitly constrained by this Task Knowledge.

\begin{itemize}
  \item \textbf{AND-Decomposition:}
        Decomposes a parent goal into a set of sub-goals, all of which must be completed.
  \item \textbf{OR-Decomposition:}
        Decomposes a parent goal into a set of optional sub-goals.
\end{itemize}

\subparagraph{\textbf{[RGR-I.3: Assign Sub-goal Executors using Environmental Knowledge]:}}
Environmental Knowledge defines the scope of what the engine can perceive and operate upon. This includes (but is not limited to) available system tools, external APIs, and the immediate execution context. The Planning Engine's output must adhere to the KAOS principle of agent Responsibility Assignment: every newly generated sub-goal must be assigned to an Agent (e.g., a specific tool, an external API, or another LLM-based agent) based on this Environmental Knowledge.

\subparagraph{\textbf{[RGR-I.4: Use Runtime Context for Dynamic Decision Adjustment]:}}
Runtime Context comprises the status and execution results of previously executed sub-goals. The Planning Engine utilizes this context to:

\begin{itemize}
  \item \textbf{Resolve dynamic parameter dependencies:}
        Use the results of prior sub-goals to instantiate subsequent sub-goals generated during the decomposition in [RGR-I.2] that depend on those results (e.g., using \textit{flight search} results to populate the parameters for \textit{book flight}).
  \item \textbf{Drive runtime error correction and re-refinement: }
        When a prior sub-goal fails, the Planning Engine must use this context—premised on the failure—to attempt a new refinement operation based on Task Knowledge and Environmental Knowledge. This is done to recover from the failure (e.g., by executing compensation tasks or backtracking) and continue execution.
\end{itemize}

\subsubsection{\textbf{Pillar II: Identification and Transformation of Runtime Requirements-Expectations}}

\begin{figure}[t]
  \centering
  \includegraphics[width=1\linewidth]{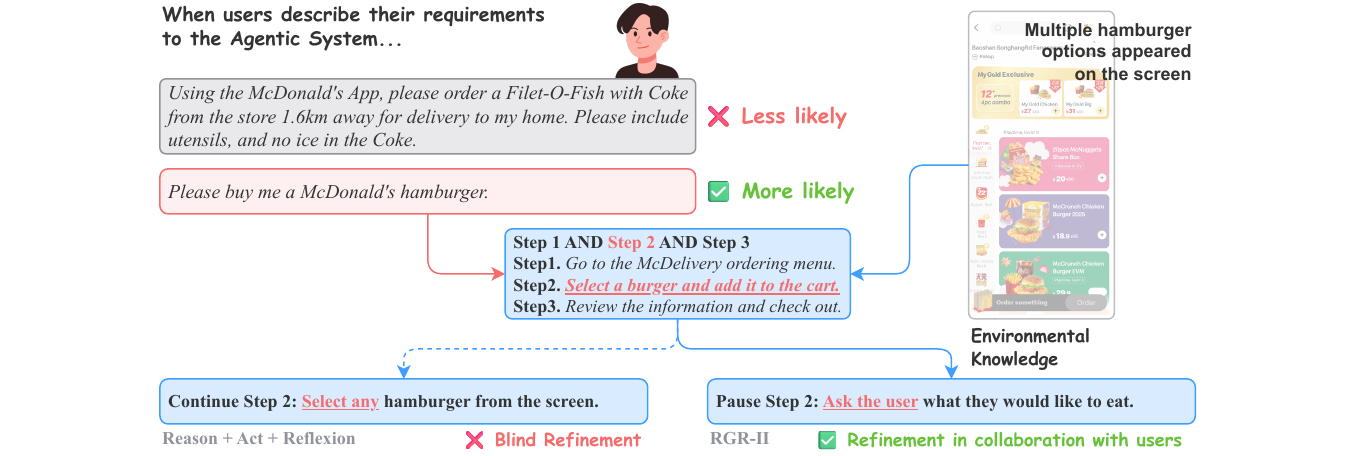}
  \caption{RGR-II (Collaborative Refinement) vs. the Traditional ReAct Paradigm (Blind Refinement). In real-world scenarios, user instructions are typically ambiguous. When the environment provides multiple feasible options (e.g., multiple hamburger options appearing on the screen that all satisfy the "buy a hamburger" request), paradigms like ReAct, being task-completion oriented, will blindly refine the user requirement and unilaterally select any hamburger, leading to a failure to meet the user's needs. RGR-II identifies this (an OR-decomposition) as a Runtime Expectation , pauses autonomous execution, and collaboratively refines with the user ("Ask the user what they would like to eat"), ensuring the execution trajectory remains aligned with the user's true intent.}
  \label{fig:rgr2}
\end{figure}

Paradigms like ReAct + Reflexion primarily address how to efficiently and robustly complete a given task, rather than how to ensure the task itself is what the user truly wants. In RGR, we reconstruct the KAOS concepts of \textbf{Requirement} and \textbf{Expectation} based on a new criterion: \textbf{whether runtime goal refinement necessitates user intervention}. We define them as two core states within a runtime closed-loop control flow and establish a user-interaction-driven mechanism for transforming one into the other. The example in Figure \ref{fig:rgr2} illustrates the starkly different behaviors exhibited when an agent system follows the ReAct + Reflexion paradigm versus the RGR-II paradigm, particularly when confronted with multiple feasible paths arising from vague user requirements. 

\paragraph{\textbf{RGR Specifications:}}

\subparagraph{\textbf{[RGR-II.1: Identify Runtime Requirements and Runtime Expectations]:}}
When refining, the Planning Engine must classify every sub-goal into one of two states:

\begin{itemize}
  \item \textbf{Runtime Requirement:}
        A clearly specified sub-goal that can be directly assigned to an agent for satisfaction (i.e., it is directly executable or can be further AND-decomposed).
  \item \textbf{Runtime Expectation:}
         A non-deterministic sub-goal that requires environmental (i.e., user) intervention to be resolved. This occurs in two scenarios: a) Knowledge Mismatch: Due to insufficient Task Knowledge or Environmental Knowledge, the goal can neither be directly executed nor further refined (e.g., lacking necessary parameters or no executable Agent can be found). b) Decision Point: The refinement results in multiple alternative paths (i.e., an OR-decomposition), which the agent cannot autonomously select.
\end{itemize}

\subparagraph{\textbf{[RGR-II.2: Refine Runtime Expectations via User Interaction]: }}
Upon identifying a Runtime Expectation, the Planning Engine must pause autonomous execution and activate User Interaction. It is strictly prohibited from autonomously selecting an OR-branch or speculating about missing information. After the user fulfills the Expectation (i.e., makes a selection or provides information via the interaction), the system must update the sub-goal based on this information. Autonomous execution resumes only after the sub-goal is transformed into a Runtime Requirement.

\subparagraph{\textbf{[RGR-II.3: Design User Interaction as an Intent Scaffolding]:}}
The User Interaction mechanism must be designed as an Intent Scaffolding. Its responsibility is to provide the user with the necessary context to fulfill the Expectation (e.g., presenting the OR-options, requesting missing parameters).

\begin{figure}[t]
  \centering
  \includegraphics[width=\linewidth]{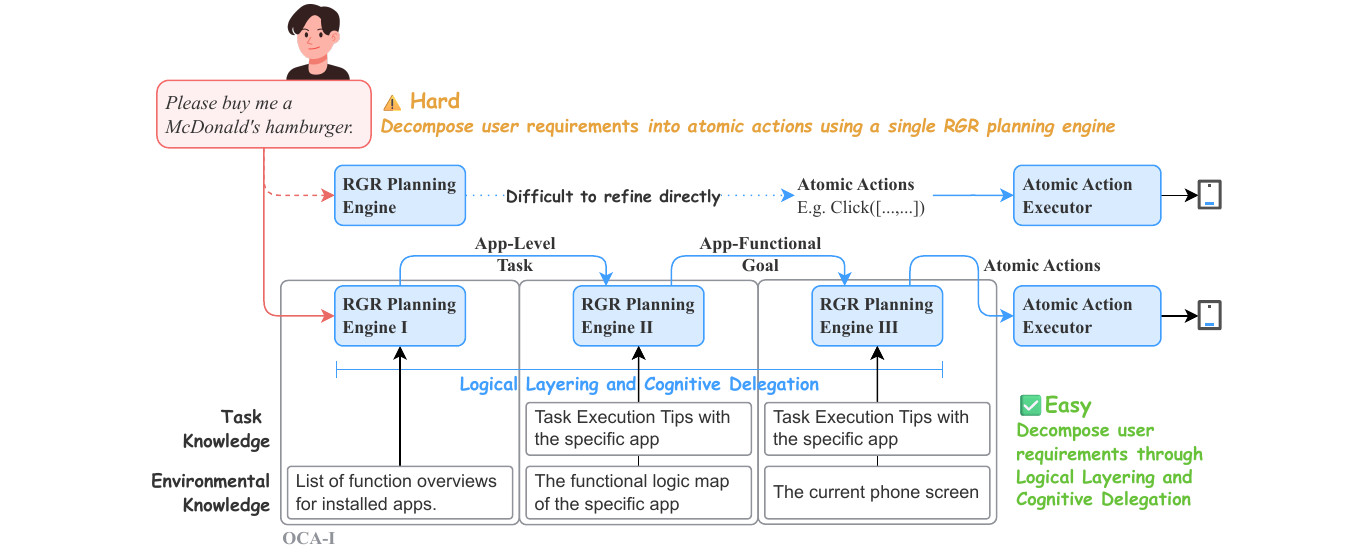}
  \caption{An architectural diagram of OCA-I (Logical Layering and Cognitive Delegation). Attempting to use a single RGR planning engine to directly decompose a high-level user intent into atomic actions is difficult (Hard), and it is also challenging to provide the appropriate Knowledge . OCA-I recommends adopting the Logical Layering and Cognitive Delegation architecture, which decomposes the task layer by layer through several RGR engines. For example: RGR Engine I is responsible for using the List of...installed apps to decompose the user intent into App-Level Tasks; RGR Engine II is responsible for refining the app-level task into App-Functional Goals ; RGR Engine III is responsible for utilizing The current phone screen"information to convert the goal into final Atomic Actions.}
  \label{fig:oca1}
\end{figure}

\subsection{The Observable Cognitive Architecture (OCA) Methodology}

Reasoning paradigms such as ReAct and Reflexion have established core behavioral models for modern agents \cite{langley2017cognitive}. However, these paradigms are essentially high-level behavioral guides, not rigorous engineering specifications. This vacuum has led to fragmented architectural implementations in practice. Developers are forced to adopt various ad-hoc methods to realize these models, resulting in a proliferation of heterogeneous, functionally-coupled, and difficult-to-observe black-box implementations. This diversity and opacity at the implementation level fundamentally undermine the observability, interpretability, and debuggability essential for complex systems.

To address this challenge, frameworks such as LangChain and AutoGen offer powerful implementations and are rapidly becoming industry de-facto standards. The contributions of these frameworks are significant. AutoGen, for instance, partially alleviates the black-box coordination and observability issues through its structured multi-agent dialogue patterns. However, as de-facto implementations, they do not constitute a complete, universal, or normative methodology. When developers adopt these tools, they are primarily adopting specific technical stacks and architectures, not following a set of general architectural principles that can guide software engineering practice.

We argue that agent engineering urgently requires a normative standard to fill this theoretical gap. To this end, we propose the Observable Cognitive Architecture (OCA) methodology. OCA aims to \textbf{provide a software engineering backbone for behavioral paradigms} like RGR and ReAct, \textbf{offering a general engineering blueprint for building observable and debuggable agent systems}.

\textbf{At the macro-logical level,} OCA adapts coordination and hierarchical concepts from Multi-Agent Systems (MAS) and Hierarchical Task Networking (HTN). It specifies \textbf{Logical Layering} and \textbf{Cognitive Delegation} to rigorously organize multiple RGR planning engines (as defined in Section 3.1). These engines progressively refine user requirements, ultimately driving the execution of reasoning paradigms like ReAct and Reflexion. \textbf{At the micro-implementation level, } OCA draws upon mature SE principles. It mandates \textbf{Cognitive Decoupling} (Single Responsibility) and \textbf{State-Control Separation} (Event-Driven Architecture). Specifically, it specifies that a Memory Bus (MB) and an Event Bus (EB) must be used to coordinate agent work, thereby achieving white-box observability. The OCA methodology is defined by specifications organized into the following two core pillars: 

\subsubsection{\textbf{Pillar I: A Multi-Agent Architecture Based on Logical Layering and Cognitive Delegation}}

Modern agent architectures lack a paradigm for organizing multi-level task planning. Furthermore, HTN, a product of the symbolic AI era, requires human experts to manually define all decomposition rules and atomic operations at design-time. In OCA, we adapt HTN's hierarchical concepts, evolving its static task decomposition into a dynamic \textbf{Cognitive Delegation} process, thereby defining—at a macro-architectural level—how multiple RGR planning engines can be integrated in a \textbf{Layered} fashion to form a reliable system. The example in Figure~\ref{fig:oca1} illustrates a multi-agent system based on the Logical Layering and Cognitive Delegation architecture described in OCA-I.

\paragraph{\textbf{OCA Specifications:}}

\subparagraph{\textbf{[OCA-I.1: Adopt Logical Layering as Integration Principle]:}
The system architecture must adopt hierarchical planning, as advocated by HTN, as its macro-integration principle. This is used to logically organize one or more planning engines (planning components in source) that conform to the RGR specifications. The system must deploy these planning engines in logical layers based on the task's abstraction level, with each layer potentially using different Task Knowledge and Environmental Knowledge.
}

\subparagraph{\textbf{[OCA-I.2: Implement Dynamic Cognitive Delegation]: }}\label{spec:oca-i-2}
HTN's decomposition method must be engineered as the agent responsibility assignment logic of a high-level RGR engine; this is termed dynamic Cognitive Delegation.

\begin{itemize}
  \item \textbf{Cognitive Delegation of Composite Tasks: }
        This agent responsibility assignment is not a mechanical task decomposition but a delegation of the cognitive process itself. A high-level engine transfers a further-refinable Runtime Requirement, along with its autonomy, to another RGR engine possessing the appropriate Task Knowledge and Environmental Knowledge. This receiving engine then initiates a new RGR cognitive process.
  \item \textbf{Delegation Pre-Validation: }
         This delegation process is non-deterministic. A high-level RGR engine must first verify that the task to be delegated is a clear Runtime Requirement. If it is not (e.g., it is a Runtime Expectation), the decomposition flow must be interrupted at that level, and the RGR User Interaction process must be executed instead.
  \item \textbf{Final Delegation of Atomic Tasks: }
         The system's architecture must define a bottom layer. RGR planning engines at this level must refine goals into Runtime Requirements that are atomic tasks. An atomic task is a final operation that requires no further cognitive delegation and can be directly assigned to an executor (e.g., a tool or API) via Environmental Knowledge.
\end{itemize}

\subsubsection{\textbf{Pillar II: Multi-Agent Collaboration via Cognitive Decoupling and State-Control Separation}}

\begin{figure}[t]
  \centering
  \includegraphics[width=\linewidth]{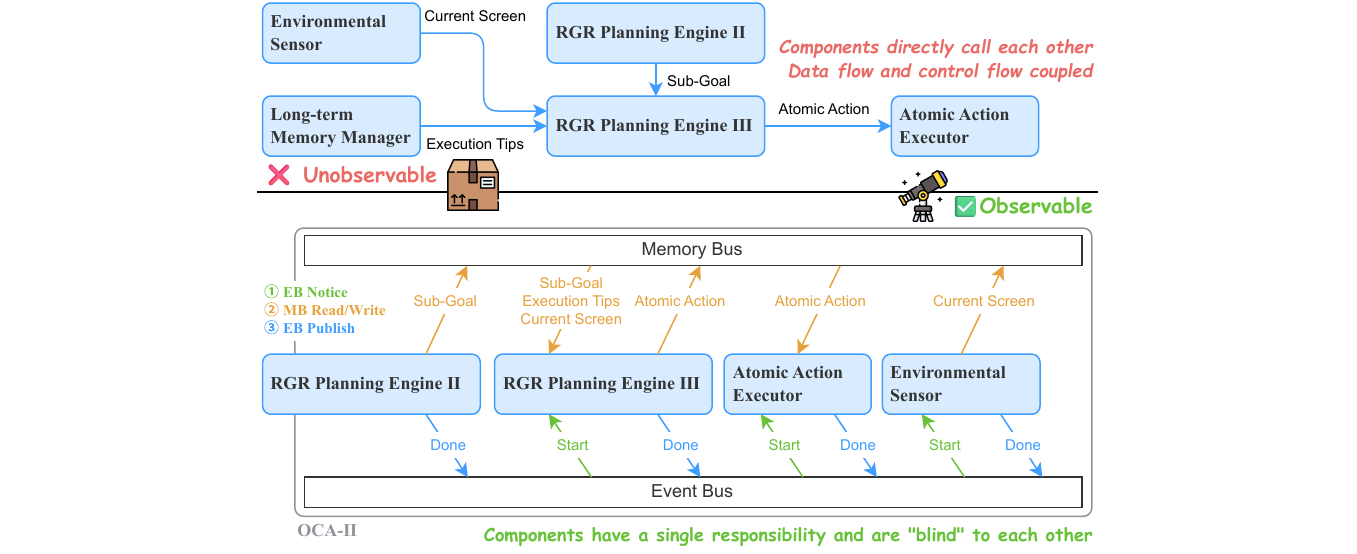}
  \caption{OCA-II (State-Control Separation) vs. Traditional Multi-Agent Black-Box Coupling. In a traditional multi-agent architecture (Top), components directly call each other, leading to severe coupling between components, which readily becomes an unobservable and difficult-to-debug black box. OCA-II achieves a white box through Cognitive Decoupling and State-Control Separation : the data flow and control flow of each component are managed by the Memory Bus (MB)and Event Bus (EB) respectively, while all components adhere to a deterministic coordination protocol, thereby ensuring system observability.}
  \label{fig:oca2}
\end{figure}

Modern reasoning paradigms like ReAct lack architectural constraints. This often leads to cognitive coupling (in single-agent setups) or opaque coordination and communication (in multi-agent setups, see Figure \ref{fig:oca2}), rendering the system an unobservable black-box. OCA fills this gap by applying \textbf{Cognitive Decoupling} and \textbf{State-Control Separation} at the micro-implementation level, structuring the multi-agent system as an observable, debuggable white-box. The example in Figure~\ref{fig:oca2} illustrates a multi-agent system implemented using the Memory Bus and Event Bus, as described in OCA-II.

\paragraph{\textbf{OCA Specifications:}}

\subparagraph{\textbf{[OCA-II.1: Implement Cognitive Decoupling Architecture]: }}
The system architecture must adopt the Cognitive Decoupling principle. An agent must not be an atomic black-box; rather, it must be engineered as an observable micro-system composed of multiple, specialized cognitive components, each with a single responsibility (e.g., planning, decision-making, reflection components). Each cognitive component should be capable of independent, lightweight model finetuning, tailored to its highly specialized task.

\subparagraph{\textbf{[OCA-II.2: Implement an Explicit Coordination State-Event Model]:  }}
The system must use a state-event hybrid model to completely separate data flow from control flow, thereby achieving white-box observability.

\begin{itemize}
  \item \textbf{Explicit State (Data Flow):  }
        The system must implement a Memory Bus (MB) to serve as the single source of truth (SSOT) for data flow and as an observable log.
  \item \textbf{Explicit Control (Control Flow):}
         The system must implement an Event Bus (EB) or an equivalent deterministic scheduler as the sole manager of control flow. Component activation must be triggered by specific events emitted from preceding components, not by data changes (i.e., not by polling the MB).
\end{itemize}

\subparagraph{\textbf{[OCA-II.3: Adhere to the Deterministic Coordination Protocol]: }}
Direct calls, message passing, or opportunistic polling of the MB among cognitive components are strictly prohibited. All components must adhere to the following deterministic protocol: Be activated by an EB event; Read state only from the MB; Write cognitive results (new state) back to the MB; Upon completion, emit a new event to the EB to trigger the next step.

\subsection{The Evolutionary Memory Architecture (EMA) Methodology}

At the forefront of AI engineering, the stateless nature of Large Language Models (LLMs) is no
longer a critical obstacle to agent development. Cognitive frameworks such as COALA \cite{Sumers2024CoALA} have provided high-level conceptual blueprints for agent mental models (e.g., memory classification) \cite{laird2012soar}.Concurrently, the emergence of dynamic RAG \cite{Su2025DynamicRAG} and persistent reflection (e.g., MPR \cite{WuQu2025MPR}) has supplied the technical and paradigmatic foundations for on-demand knowledge retrieval and the transformation of experiences into long-term rules. However, this field faces a dilemma similar to that of the ReAct paradigm: a lack of specific constraints and normative guidance for engineering implementation. This results in the absence of a unified, formal process model to orchestrate complex memory update activities. The critical memory consolidation stage remains an ad-hoc engineering effort, lacking reusable process specifications. Although technical platforms like MemOS have recently shown promise, as we emphasized in Section 3.2, specific toolkits do not provide a complete methodology for designing long-term, evolvable agent systems.

To preliminarily address this challenge, we draw upon mature concepts from autonomous computing in
traditional software engineering, particularly the MAPE-K control loop \cite{delemos2013selfadaptive2}. We propose the EMA framework as an engineering specification that bridges high-level cognitive paradigms (like COALA) with low-level technical implementations. It aims to \textbf{provide a viable memory architecture reference for building evolvable, high-reliability agent systems}. EMA provides engineering constraints along two core dimensions: \textbf{1) Standardizing How Knowledge is Stored}: It defines a hierarchical structure comprising Working Memory and Long-Term Memory, along with read/write control protocols.\textbf{2) Standardizing Where Knowledge Comes From}: It proposes a decoupled Execution-Evolution dual-loop model, designed to separate real-time tactical adaptation from asynchronous strategic evolution at an engineering level. The EMA methodology is defined by the following two core pillars: 

\begin{figure}[t]
  \centering
  \includegraphics[width=\linewidth]{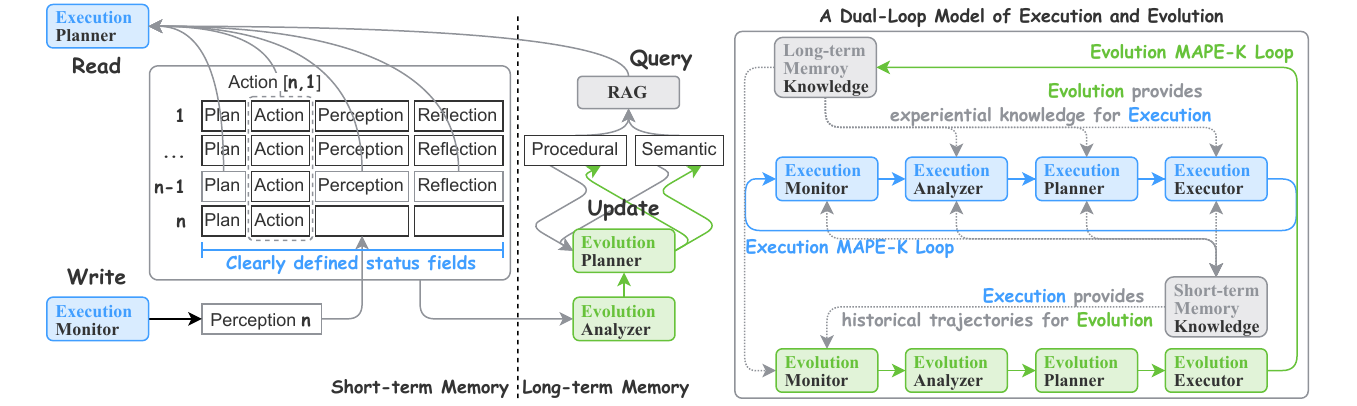}
  \caption{EMA (Evolutionary Memory Architecture)'s Hierarchical Memory Structure and Execution-Evolution Dual-Loop Model. EMA-I defines how knowledge is stored: Working Memory is a hierarchical Cognitive Stack, where agents in the Execution Loop write status to the stack top or read status from it. Long-Term Memory is divided into Procedural and Semantic knowledge . It is queried by agents during the Execution Loop at runtime, and updated by the Evolution Loop after the Execution Loop concludes, utilizing the complete trajectory from the Cognitive Stack. EMA-II further addresses how knowledge is used and where it comes from: The Execution Loop corrects tactical deviations caused by a non-deterministic environment via a runtime reflection mechanism; the Evolution Loop consolidates raw experiences from Working Memory into reusable capabilities within Long-Term Memory.}
  \label{fig:ema}
\end{figure}

\subsubsection{\textbf{Pillar I: Explicit Memory Structure and Access Control}}

In modern cognitive blueprints such as COALA, memory hierarchy is a high-level conceptual model that lacks formal specification. EMA addresses this by defining concrete \textbf{Internal Structures} for Working Memory and Long-Term Memory and specifying \textbf{Read/Write Control Protocols}. This standardizes how knowledge is stored, providing the foundation for in-task execution and post-task evolution. The example in Figure~\ref{fig:ema} illustrates the physical structure and access methods for Working Memory and Long-Term Memory as described in EMA-I.

\paragraph{\textbf{EMA Specifications:}}

\subparagraph{\textbf{[EMA-I.1: Implement Working Memory]: }}
The system must implement a Working Memory module. Its core is a hierarchical Cognitive Stack used to record the complete trajectory of the Execution Loop with high fidelity. Clear state fields must be defined for this Working Memory (e.g., a state tuple containing {Plan, Action, Perception, Reflection}). The contents of this memory are treated as volatile; regardless of whether it is persisted, subsequent tasks must not directly read or write the original instance. After task completion, its complete trajectory (full stack) serves as the analysis input for the Evolution Loop.

\subparagraph{\textbf{[EMA-I.2: Implement Long-Term Memory]: }}
The system must implement a Long-Term Memory module to persistently store knowledge abstracted and consolidated from experience during the Evolution Loop. Its internal structure must distinguish between different knowledge types, including at least the following:

\begin{itemize}
  \item \textbf{Procedural Memory: }
        Stores skills and strategies regarding how to do. This manifests physically as operational techniques that can be retrieved and reused based on task similarity, providing Task Knowledge for planning, execution, and error recovery.
  \item \textbf{Semantic Memory: }
         Stores declarative knowledge about what the environment is. This manifests physically as environment maps or knowledge graphs that can be retrieved and reused based on the current context, providing Environmental Knowledge for task decomposition and planning.
\end{itemize}

\subparagraph{\textbf{[EMA-I.3: Implement Memory Access Control]: }}
The system must control and separate the read/write operations for different memory modules:

\noindent \textbf{1) Working Memory: }
  \begin{itemize}
    \item \textbf{Write:}
        During the Execution Loop, cognitive components (e.g., planner, executor, perceiver) perform high-frequency, layered writes by pushing a new state onto the Cognitive Stack.
    \item \textbf{Read: }
         Must support differentiated reading. During the Execution Loop, cognitive components can read the stack top for the current state or retrieve specific layers for recent history. After the task ends, the Evolution Loop must read the full stack (complete trajectory) for analysis.
  \end{itemize}

\noindent \textbf{2) Long-Term Memory: }
  \begin{itemize}
    \item \textbf{Update:}
        Must be protected as read-only during the Execution Loop. Only the Evolution Loop is permitted to perform update operations (add, modify, delete) after the task is complete.
    \item \textbf{Query:}
         During the Execution Loop, it is queried (e.g., via RAG\cite{Lewis2020RAG}) by planning and decision components (e.g., planner, action decider) to guide their execution.
  \end{itemize}       

\noindent This specification serves as one possible implementation of the Memory Bus defined in [OCA-II.2].

\subsubsection{\textbf{Pillar II: The Execution-Evolution Dual-Loop Model}}

In the classic MAPE-K model, the control loop is a monolithic process for system adaptation (e.g., parameter tuning), and its Knowledge Base (K) is treated as a static resource. EMA reconstructs this single MAPE-K loop, shifting its mission from external tuning to internal evolution (i.e., the evolution of K itself). We decouple it into a dual-loop model: an Execution Loop for rapid, in-task tactical adaptation, and an Evolution Loop for accumulating knowledge for long-term strategic evolution. EMA thus standardizes where knowledge comes from, providing the foundation for the knowledge required by [RGR-I.2] and [RGR-I.4]. The example in Figure~\ref{fig:ema} illustrates the dual-loop process of the Execution Loop and Evolution Loop described in EMA-II.

\paragraph{\textbf{EMA Specifications:}}

\subparagraph{\textbf{[EMA-II.1: Implement Execution Loop]:}}
The system must implement a fast-response, in-loop MAPE-K cycle, designated as the Execution Loop. Its goal is to ensure task success by correcting tactical deviations caused by a non-deterministic environment, using a runtime reflection mechanism. It achieves this tactical adaptation by continuously: Monitoring action results, Analyzing deviations, Planning corrective actions, and Executing those corrections. In this loop, the Knowledge Base (K) is used with high frequency: it reads/writes Working Memory as cognitive scaffolding and reads Long-Term Memory for decision guidance.

\subparagraph{\textbf{[EMA-II.2: Implement Evolution Loop]: }}
The system must implement a slow, post-hoc MAPE-K cycle, designated as the Evolution Loop. Its goal is to drive system evolution by consolidating raw experience from Working Memory into reusable capabilities within Long-Term Memory. It is defined as follows:

\begin{itemize}
  \item \textbf{M (Monitor):}
         Read the complete task trajectory (from Working Memory) and the currently accumulated Long-Term Memory.
  \item \textbf{A (Analyze): }
         Conduct high-level reflection on the full trajectory to identify patterned failures, redundant steps, or more efficient strategies.
  \item \textbf{P (Plan):}
         Generate a knowledge update plan, i.e., plan how to abstract newly learned experiences into Procedural Memory and Semantic Memory.
  \item \textbf{E (Execute):}
         Execute the knowledge update plan, writing or overwriting the new knowledge into Long-Term Memory (K).
  \item \textbf{K (Knowledge): }
         In this loop, the Long-Term Memory (K) is the final output and target of the update itself.
\end{itemize}

\section{Case Study: Fairy Mobile GUI Agent}
\label{ch:case_study_fairy}

\begin{figure}[t]
  \centering
  \includegraphics[width=\linewidth]{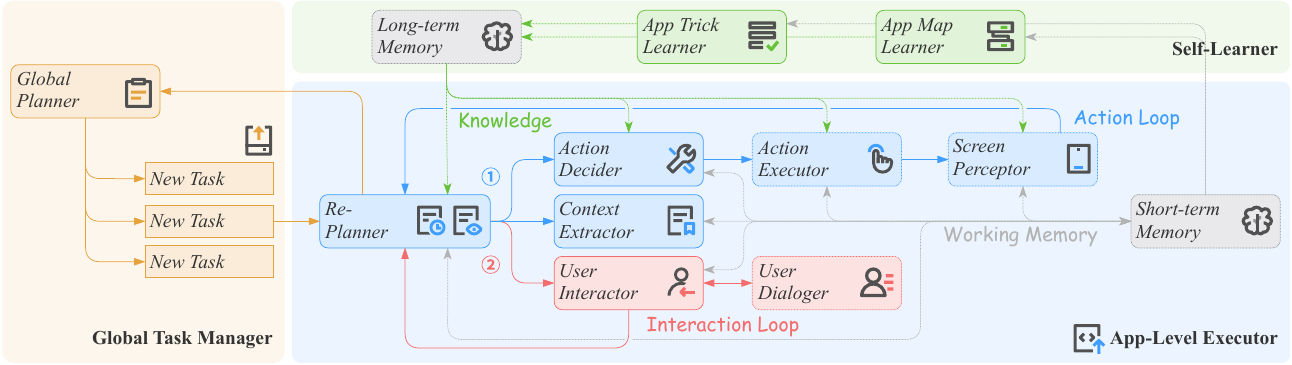}
  \caption{Overview of the Fairy framework. Fairy consists of three core components: the Global Task Manager, the App-Level Executor, and the Self-Learner. When a user issues a task instruction, the Global Task Manager decomposes the task into app-specific sub-tasks and delegates them to the App-Level Executor for execution. The App-Level Executor, guided by the Re-Planner, utilizes two key loops—the Action Loop and the Interaction Loop—to continuously execute the task and interact with the user. After task completion, the Self-Learner summarizes knowledge from the task and stores it as long-term memory for future use.}
  \label{fig: ov}
\end{figure}

In Chapter \ref{ch:engineering_framework}, we proposed a systematic software engineering framework composed of RGR, OCA, and EMA. This framework aims to provide a structured specification for building robust, maintainable, and evolvable Agentic AI systems. However, the engineering value of a theoretical framework—including its improvements to task performance, system maintainability, and scalability—must be empirically validated by instantiating it within a concrete, non-trivial system.

\textbf{We chose the domain of Mobile GUI Agents to validate our framework.} This is because the existing challenges in this domain align closely with the core problems our framework aims to solve, and these challenges also epitomize the typical difficulties faced by current Agentic systems. Furthermore, the limitations in the engineering implementation of SoTA methods in this field provide an ideal control group for our framework. This domain faces the following core challenges:

\begin{itemize}
    \item \textbf{Vague Instruction Challenge:} As mobile assistants, agents directly receive commands from non-expert users, inevitably facing instruction ambiguity. Users rarely provide all requirements accurately in a single instance. This necessitates the agent's ability to interactively communicate with the user during execution to clarify intent.

    \item \textbf{Complex Task Challenge:} In real-world scenarios, mobile tasks may involve multi-stage, cross-app execution. Concurrently, the functionality of apps themselves is increasingly complex, posing significant challenges in task planning and error recovery. This requires the agent to possess strong task decomposition capabilities and the ability to continuously track and adjust plans during execution.

    \item \textbf{Experience Accumulation Challenge:} The mobile app ecosystem exhibits a significant long-tail effect, and app interfaces and functionalities are frequently updated. It is difficult for any agent to rely solely on pre-trained commonsense to handle this diversity and volatility. This requires the agent to possess the ability to accumulate experience and self-evolve during usage.
\end{itemize}

As an emerging and active research field, Mobile GUI Agents have seen numerous works attempting to overcome the aforementioned challenges from various perspectives. \textbf{However, many existing SoTA works have evolved independently from an AI-first perspective. In terms of architectural design, they do not follow systematic engineering specifications.} Instead, they have sporadically applied and re-invented software engineering concepts. This ad-hoc design not only directly causes current agent systems to devolve into black boxes that are difficult to observe, debug, and extend, but also indirectly results in their failure to completely overcome the above challenges, thereby hindering the final application of these works.

We argue that the pain points exposed by these SoTA methods are, in essence, a lack of software engineering. Therefore, in this chapter, we will apply the systematic software engineering framework from Chapter \ref{ch:engineering_framework} to construct an entirely new mobile GUI agent system, Fairy. We will subsequently evaluate it to validate the effectiveness of this software engineering framework.

\subsection{Fairy: An Interactive and Self-Evolving Mobile Assistant based on Multi-Agent}
\label{sec:fairy_intro}

We have developed Fairy, a mobile GUI agent that fully implements the theoretical specifications from Chapter 3 in its architecture. It is designed as an interactive, self-evolving multi-agent system, and its overall architecture is composed of three core modules:

\begin{itemize}
    \item \textbf{Global Task Manager}: As the top-level coordinator, this module is responsible for decomposing the user's instruction into a series of sub-tasks specific to individual apps.
    \item \textbf{App-Level Executor}: A core execution unit composed of multiple agents. It handles in-app planning, execution, and user interaction through an Action Loop and an Interaction Loop.
    \item \textbf{Self-Learner}: Responsible for post-execution reflection, consolidating experience (Working Memory) into reusable long-term knowledge (specifically, App Map and App Tricks).
\end{itemize}

In the subsequent sections of this chapter, we will detail how these specific architectural components of Fairy serve as precise engineering instantiations of the RGR, OCA, and EMA specifications, respectively. This provides a measurable implementation basis for the empirical evaluations in Chapters \ref{ch:Evaluation} and \ref{ch:Result&Discussion}.

Fairy have been open-sourced on GitHub: https://github.com/NeoSunJZ/Fairy/

\subsection{Macro-level Logical Architecture of the OCA Specification}
\label{sec:oca_macro_architecture}

The essence of Agentic Software Engineering is to build systems capable of autonomously following a user's high-level intent and executing complex tasks. Therefore, a plan-execute model is the necessary architectural starting point for building such systems. In the Mobile GUI Agent domain, this model becomes the core around which architectural design revolves: the user's high-level instruction (e.g., \textit{Help me order a McDonald's burger}) must be transformed into a series of atomic operations executable on the device (e.g., Tap(x,y), Input(...)).

However, it is precisely in this transformation that the system faces a fundamental engineering challenge: the vast semantic gap between user high-level intent and device atomic operations. A single, end-to-end, monolithic agent will inevitably fail when attempting to bridge this gap in one go, due to overly long reasoning chains and state-space explosion. Therefore, adopting the fundamental engineering principle of Decomposition—refining this gap layer by layer—becomes an engineering necessity. This is the core idea advocated by [OCA-I], which provides systematic architectural guidance for solving such complexity challenges.

Under the guidance of [OCA-I], Fairy's architecture naturally evolves this decomposition process into a three-layer refinement structure:

\begin{enumerate}
    \item \textbf{First Decomposition: From User Intent to App-Level Tasks}: In real mobile scenarios, user intent often has cross-app characteristics (e.g., "find a nearby well-rated restaurant and organize it into a memo"). Therefore, a natural and logical first decomposition is to break down the user's single high-level intent into sub-tasks executed on different applications.

    \item \textbf{Second Decomposition: From App-Level Task to Sub-Goals}: An in-app task (e.g., "find a nearby well-rated restaurant") is still highly complex, and its granularity remains significantly different from the final atomic operations. A second decomposition is necessary: this layer's responsibility is to orchestrate around the app's core functional points, further decomposing the macro-level in-app task into a sequence of logically clear sub-goals (e.g., [SG1: Search for burger, SG2: Select combo, SG3: Confirm payment]).

    \item \textbf{Third Decomposition: From Sub-Goal to Atomic Operations}: Even after two decompositions, a sub-goal is still not a directly executable operation. Therefore, we must perform a third decomposition, further breaking down the sub-goal (e.g., "Search for burger") into a sequence of atomic operations (e.g., Tap(search\_box), ClearInput(), Input('burger'), Tap(search\_button)).
\end{enumerate}

In summary, Fairy's Global-App-Atomic three-layer architecture is the necessary product of solving the core engineering challenge of the semantic gap. It precisely instantiates the logical layering defined by [OCA-I].

However, this macro-level architecture only defines the structure of the decomposition. We still need to define the cognitive components that execute this decomposition within the structure. This is precisely the responsibility of the RGR—it defines the Planning Engine responsible for performing goal refinement at runtime. Therefore, in the following Section \ref{sec:rgr_planning_engines}, we will delve into Fairy's internal architecture to detail how Fairy instantiates these Planning Engine components at each of the three decomposition layers, and demonstrate how they implement [RGR-I] and [RGR-II].

\subsection{Planning Engine Components of the RGR Specification}
\label{sec:rgr_planning_engines}

In this section, based on the macro-level architecture of Fairy established in Section \ref{sec:oca_macro_architecture}, we will define the Planning Engines that actually execute goal decomposition within this structure. [Specification RGR-I] imposes three strict engineering requirements on a Planning Engine: it must, at runtime, explicitly rely on the input of Task Knowledge, Environmental Knowledge, and Runtime Context, rather than depending on the model's commonsense for black-box speculation. This specification has a direct implication for the system architecture: every Planning Engine component within Fairy's three-layer architecture must be engineered as a complete cognitive unit, which should include:

\begin{itemize}
    \item \textbf{Core Logic}: The agent ($\mathcal{A}$) itself, responsible for performing task decomposition based on knowledge.
    \item \textbf{Sensor}: A tool or mechanism for acquiring Environmental Knowledge.
    \item \textbf{Executor / Delegator}: A mechanism for either executing atomic operations (as defined by [OCA-I]) or cognitively delegating the refined sub-goals to the next-level planning engine.
\end{itemize}

In the following subsections, we will demonstrate, layer by layer, how Fairy's various Planning Engine components implement [Specification RGR-I], and we will focus our discussion on Fairy's specific design for [RGR-II].

\subsubsection{High-Level Planning Engine.}
\label{sec:high_level_planner}

\begin{figure}[t]
  \centering
  \includegraphics[width=\linewidth]{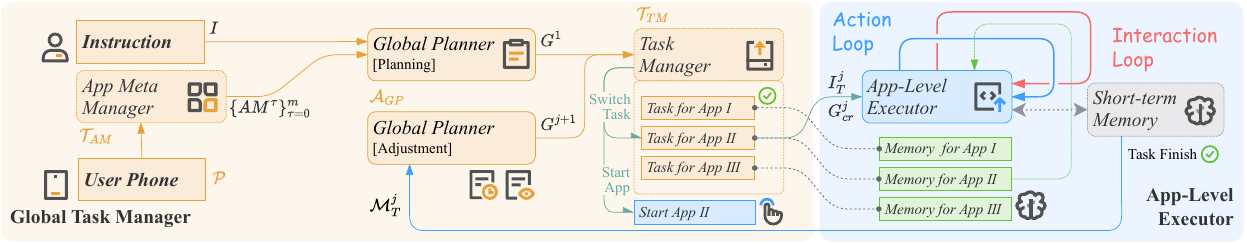}
  \caption{The primary workflow of the Global Task Manager. As the top-level task planning module, the Global Task Manager includes the Global Re-Planner agent, responsible for task decomposition and reflection at the global level. It also includes the App Info Manager and Task Manager tools, responsible for managing the metadata of installed applications on the device and for dispatching tasks to the App-Level Executor and managing their lifecycle, respectively. The App-Level Executor loads different memories according to the target application, thereby exhibiting differentiated task execution capabilities.}
  \label{fig:GlobalTaskManager}
\end{figure}

Fairy's high-level planning engine is the core component for decomposing user instructions into a series of app-level tasks. As shown in Figure \ref{fig:GlobalTaskManager}, this planning engine is composed of the core planner Global Planner, the sensor App Metadata Manager, and the delegation-assisting tool Task Manager. In the Fairy system, this entire unit is named the Global Task Manager.

\noindent \textbf{1) Core Planner:} The Global Planner ($\mathcal{A}_{GP}$) is the core of the high-level planning engine. It is responsible for decomposing the user's complex instruction into multiple tasks that can be executed individually on apps installed on the user's phone. This process is divided into two stages: Direct Planning and Adjustment:

\begin{enumerate}
    \item \textbf{Direct Planning:} $\mathcal{A}_{GP}$ constructs a global plan $G$ based on the User Instruction $I$ and the Installed Apps Metadata $\{AM^\tau\}^ m_{\tau=0}$, and determines the first sub-task to be executed.
$$G^1=\mathcal{A}_{GP}(I,\{AM^\tau\}^ m_{\tau=0})$$
$G$ consists of the global overall plan $G_{st}$, the sub-task $G_{st}$ to be executed next, and the contextual information $G_{c}$ required for executing this sub-task. The sub-task $G_{st}$ includes the raw task instruction $I_{T'}$, the context extraction request $G_{cr}$, and the target app package name $AM_{p}$.

    \item \textbf{Adjustment:} $\mathcal{A}_{GP}$ performs (a) inspecting previous execution results, and (b) updating task progress or revising the plan. $\mathcal{A}_{GP}$ determines whether the task result meets expectations based on the user instruction $I$ and the execution trace $\mathcal{M}_{T}$ generated by the App-Level Executor through multiple rounds of action and user-interaction cycles. Then, based on this evaluation and the previous global plan $G$, it updates the $G_{op}$, identifies the next $G_{st}$, and summarizes the $G_{c}$ to be passed to the next sub-task.
$$G^{j+1}=\mathcal{A}_{GP}(I,\mathcal{M}_{T}^j,G^{j})$$

\end{enumerate}

\noindent \textbf{2) Sensor}: The App Metadata Manager ($\mathcal{T}_{AM}$) is a sensor (tool) for discovering, maintaining, and summarizing the metadata of apps installed on the user's device $\mathcal{P}$, providing $\mathcal{A}_{GP}$ with the necessary environmental context for delegation (i.e., which apps are on the phone and what functions they have). This information is dynamically updated as the installed apps change.
$$\{AM^\tau_{p},AM^\tau_{d}\}^m_{\tau=0}=\mathcal{T}_{AM}(\mathcal{T}_{AE}.ListApp(\mathcal{P}))$$

\noindent \textbf{3) Cognitive Delegation:} The Task Manager ($\mathcal{T}_{TM}$) is a tool to assist cognitive delegation, managing the lifecycle of sub-tasks planned by $\mathcal{A}_{GRP}$ and assisting in context transfer during task switches. Upon receiving the current $G$, $\mathcal{T}_{TM}$ calls $\mathcal{T}_{AE}.StartApp(\mathcal{P},AM_{p})$ to launch the target app. Concurrently, $\mathcal{T}_{TM}$ uses an LLM to rewrite the raw task instruction $I_{T'}$ based on the contextual information $G_{c}$. After the task is issued, the task instruction $I^j_{T}$ will be directly delegated to the mid-level planning engine $\mathcal{A}_{RP}$ (described in Section \ref{sec:middle_level_planner}) for further processing.
$$I^j_{T} = \mathcal{T}_{TM}(I^j_{T'}, G_{c}^{j})$$

\subsubsection{Middle-Level planning engine.}
\label{sec:middle_level_planner}

\begin{figure}[t]
  \centering
  \includegraphics[width=\linewidth]{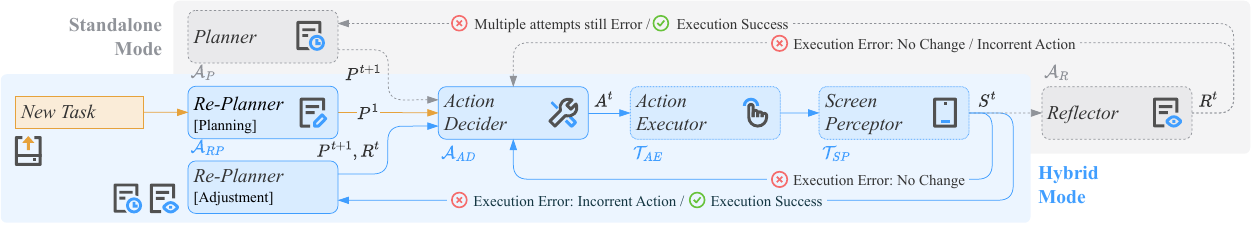}
  \caption{The primary workflow of the Action Loop. In the Action Loop, the Re-Planner and Action Decider agents are responsible for mid-level task planning and reflection, and for deciding the specific actions to execute at the low level. Concurrently, the Action Executor and Screen Perceptor tools are responsible for executing actions and perceiving screen information after execution to track task progress. Furthermore, the Key Context Extractor agent continuously collects key contextual information from the screen within the Action Loop to ensure that important information is effectively transmitted during the task execution.}
  \label{fig:ActionLoop}
\end{figure}

Fairy's mid-level planning engine is the core component for decomposing app-level tasks into a series of functional sub-goals. As shown in Figure \ref{fig:ActionLoop}, this planning engine refers to the Re-Planner. It requires no additional environmental sensing, as it does not assign sub-goal executors. Similarly, it lacks a direct cognitive delegation tool, as the lower-level engine requires no special prerequisite actions before decomposition. Due to its app-internal planning scope and its results being directly consumed by the low-level engine, it is considered a component of the App-Level Executor module within the Fairy system.

\noindent \textbf{1) Core Planner:} Re-Planner ($\mathcal{A}_{RP}$) is the core of the mid-level planning engine, responsible for generating a concrete plan with sub-goals based on the task instruction $I_{T}$ from ($\mathcal{A}_{RP}$). Similar to the Global Planner, $\mathcal{A}_{RP}$ also consists of two stages:
\begin{enumerate}
    \item \textbf{Direct Planning:} $\mathcal{A}_{RP}$ constructs a plan $P$ based on the task instruction $I_T$, first screen perception $S^0$, and planning tricks $T^{I_T}_p$, and determines the first sub-goal to be executed.
    $$P^1=\mathcal{A}_{RP}(I_T,S^{0},T^{I_T}_{p})$$
    $P$ consists of the overall plan $P_{Op}$ and the sub-goal $P_{Sg}$ to be executed next.

    \item \textbf{Adjustment:} $\mathcal{A}_{RP}$ performs (a) inspecting previous action execution results, and (b) updating task progress or revising the plan. First, $\mathcal{A}_{RP}$ inspects whether the screen change meets the action expectation, given the task instruction $I_T$, previous screen perception $S$, and the action execution trace $\mathcal{M}_{A}$, and tracks the progress of the current sub-goal $P_{Sg}$, outputting a reflection result $R$. Next, based on the reflection result $R$, the previous plan $P$, and key context $C$, $\mathcal{A}_{RP}$ updates the $P_{Op}$ and determines the next $P_{Sg}$. Finally, $\mathcal{A}_{RP}$ decides whether to interact with the user and outputs an interaction request $D_R$.
    $$P^{t+1}, D^{t+1}_R,R^{t}=\mathcal{A}_{RP}(I_T,S^{t-1},\mathcal{M}_A^t,C^{t-1},T^{I_T}_{p})$$
    $D_{R}$ includes the interaction type $D_{Ic}$ and thought process $D_{It}$; $R$ includes the action result $R_{Ar}$ and plan progress $R_{Pr}$. The possible values of $R_{Ar}$ are: A: sub-goal completed; B: sub-goal partially completed; C: unexpected outcome; D: no screen change. When $R_{Ar}$ in C, D , $R$ also includes potential error causes $R_{Ec}$ for recovery. When $R_{Ar}$ is A, $\mathcal{A}_{RP}$ selects the next sub-goal from $P_{Op}$; otherwise, the current sub-goal remains unchanged. When $R_{Ar}$ is C or D for three consecutive times, $\mathcal{A}_{RP}$ considers revising $P_{Op}$ and selecting a new sub-goal, while keeping the already completed parts unchanged.
    $\mathcal{A}_{RP}$ supports the Standalone mode, where duties (a) and (b) are split between the Reflector ($\mathcal{A}_{R}$) and the Planner ($\mathcal{A}_{P}$). This mode slightly improves stability and accuracy at the cost of slower execution.
\end{enumerate}

\noindent \textbf{2) Key Context Collector:} The Context Extractor ($\mathcal{A}_{CE}$) extracts key contextual information from the screen for use by both $\mathcal{A}_{RP}$ and $\mathcal{A}_{AD}$. In information-gathering tasks (e.g., finding nearby restaurants and summarizing reviews), the context maintained by $\mathcal{A}_{CE}$ is essential. Given the task instruction $I_T$, context extraction request $G_{cr}$, plan $P$, current screen perception $S$, and previous context $C^{t-1}$, it produces a new key context $C^t$:
$$C^{t}=\mathcal{A}_{CE}(I_{T},G_{cr},P^t,S^{t},C^{t-1})$$
After each round of the Action Loop, $\mathcal{A}_{CE}$ is activated only when $R_{Ar}$ in A,B. $\mathcal{A}_{CE}$ operates in parallel with other agents, which do not need to wait for it to finish collecting information, as the initial screen perception of the current round already contains information from the previous one. This design improves the agent's response speed.

\noindent \textbf{3) Cognitive Delegation:} The output $P$ of $\mathcal{A}_{RP}$ is directly delegated to the low-level Planning Engine, $\mathcal{A}_{AD}$ (described in Section \ref{sec:low_level_planner}), for further processing.

\subsubsection{Low-Level planning engine.}
\label{sec:low_level_planner}

Fairy's low-level Planning Engine is the core component for decomposing functional sub-goals into a series of atomic operations directly executable on the mobile device. As shown in Figure \ref{fig:GlobalTaskManager}, this Planning Engine comprises the core planner Action Decider, the sensor Screen Perceptor, and the executor Action Executor. In the Fairy system, this unit is considered a core part of the App-Level Executor module.

\noindent \textbf{1) Core Planner:} The Action Decider ($\mathcal{A}_{AD}$) is the core of the low-level planning engine, responsible for translating the sub-goal planned by $\mathcal{A}_{RP}$ into concrete atomic actions. Based on the previous action result $R_{Ar}$, it first retrieves either the action decision tricks $T^{P_{Sg}}_{exe}$ or the error recovery tricks $T^{R_{Ec}}_{err}$. Then, given the task instruction $I_T$, current plan $P$, screen perception $S$, recent n execution records $\{ \mathcal{M}_{A}^{\tau} \}_{\tau = t-n}^{t}$, and key context $C$, it generates an atomic action sequence and outputs the action decision $A$.
$$A^t=\mathcal{A}_{AD}(I_{T},P^t,S^t, \{ \mathcal{M}_{A}^{\tau} \}_{\tau = t-n}^{t},C^t,T^{P_{Sg}}_{exe} / T^{R_{Ec}}_{err})$$
$A$ includes the atomic action sequence $A_{As}$ and the expected result $A_{Er}$. $A_{As}$ consists of one or more atomic operations, with parameters included if required.

\begin{figure}[t]
  \centering
  \includegraphics[width=\linewidth]{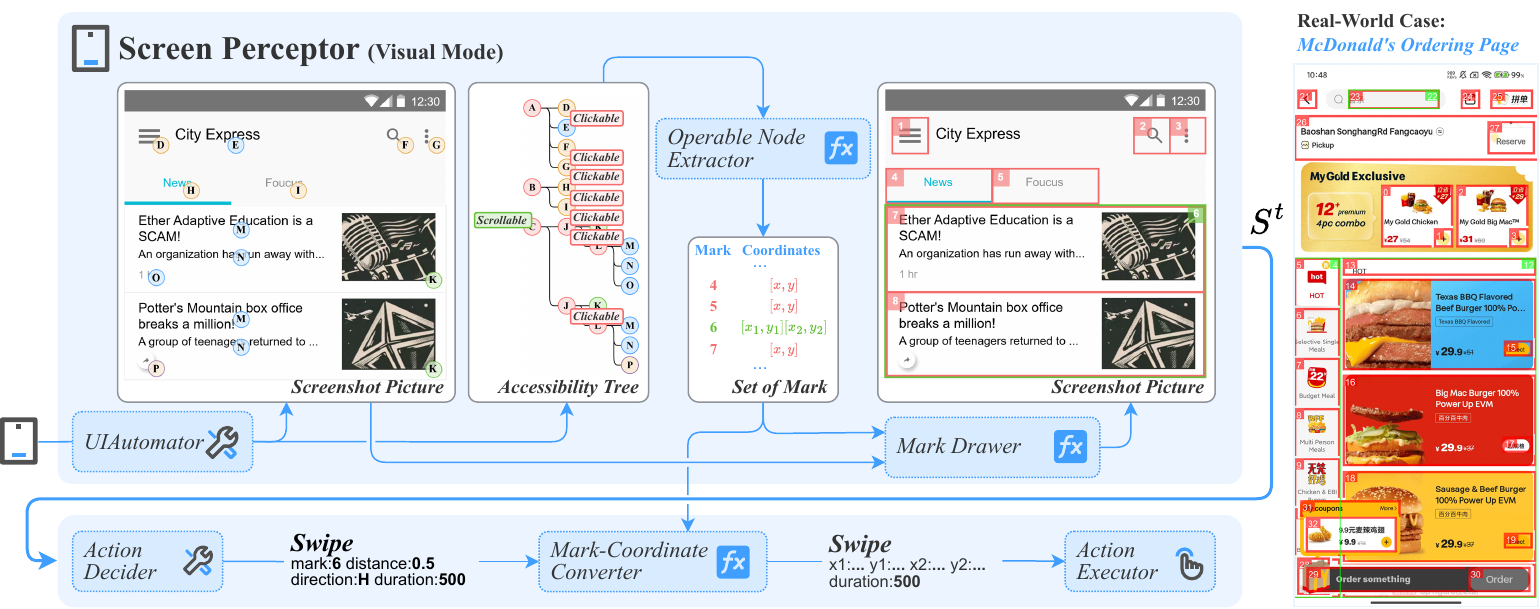}
  \caption{Overview of the Screen Perceptor (Visual vode) workflow, The module collects screenshots and the AccessibilitTree, identifies operable nodes, assigns visual marks, and renders bounding boxes for action decision-making. As shown in thereal-world McDonald's ordering page, obscured components (e.g., sidebar under a tooltip) are automatically invalidated andexcluded from decision-making.}
  \label{fig:Screen-Perceptor}
\end{figure}

\noindent \textbf{2) Sensor: } The Screen Perceptor ($\mathcal{T}_{SP}$) is a sensor (tool) for capturing and processing the current screen state, providing $\mathcal{A}_{AD}$ with the necessary environmental context for delegation (i.e., what components are on the screen and where the target component is). $\mathcal{T}_{SP}$ supports capturing both a screenshot $S_{Sp}$ and the accessibility tree $S_{At}$ (in XML format) via UI Automator\cite{Wang2016UIAutomator}. It supports two modes:

\begin{enumerate}
    \item \textbf{Visual Mode}: As shown in Figure \ref{fig:Screen-Perceptor}, $\mathcal{T}_{SP}$ extracts clickable and scrollable components from $S_{At}$, uses their bounds to locate and mark them on $S_{Sp}$, and assigns a mark index. The mark-to-coordinate mapping (Set of Mark, SoM) is stored as $S_{SoM}$. In this mode, $A_{As}$ is converted based on $S_{SoM}$ before execution by $\mathcal{T}_{AE}$.
    \item \textbf{Non-visual Mode}: $\mathcal{T}_{SP}$ prunes irrelevant nodes and attributes from $S_{At}$, completes image node descriptions using an MLLM, merges clickable child nodes, and outputs a structured Markdown-like representation $S_{Pt}$ as textual perception. This enables support for non-visual models.
\end{enumerate}

\noindent \textbf{3) Target Executor:} Action Executor ($\mathcal{T}_{AE}$) is a tool, used for finally executing the atomic action sequence $A_{As}$ generated by $\mathcal{A}_{AD}$. $\mathcal{T}_{AE}$ can, via UIAutomator, execute the following atomic operations on the user device $\mathcal{P}$:

\begin{itemize}
    \item \textbf{Tap(x, y)}: Tap at coordinate (x, y) on the current screen.
    \item \textbf{Swipe(x1, y1, x2, y2, duration)}: Swipe from coordinate (x1, y1) to (x2, y2), duration is duration seconds.
    \item \textbf{LongPress(x, y, duration)}: Long-press at (x, y), duration is duration seconds.
    \item \textbf{Input(text)}: Types text into the active input field.
    \item \textbf{ClearInput()}: Clear the content in the current input field.
    \item \textbf{KeyEvent(type)}: Send a key event, type corresponds to Android key code, e.g., KEYCODE\_BACK or KEYCODE\_HOME.
    \item \textbf{Wait(duration)}: Wait for duration seconds.
    \item \textbf{Finish()}: Task completion mark, performs no action.
    \item \textbf{NeedInteraction()}: Signals that user interaction is required; performs no action.
    \item \textbf{ListApps()}: Returns the package names of all third-party apps installed on the device.
    \item \textbf{StartApp(app\_package\_name)}: Launches the specified application.
\end{itemize}

\subsubsection{Runtime Requirement Discovery and User Interaction.}
\label{sec:runtime_discovery_interaction}

Sections \ref{sec:high_level_planner} to \ref{sec:low_level_planner} jointly defined the Fairy system's top-down pipeline for refining user instructions and executing autonomously, a process also known as the Action Loop in Fairy's architecture. The Planning Engines within this Action Loop adhere to [RGR-I], solving the core engineering problem of decomposing a well-defined user instruction into a series of executable atomic operations. However, according to [RGR-II], these Planning Engines must also identify runtime requirements and runtime expectations, and use User Interaction to refine those expectations.

A key engineering decision was how to implement [RGR-II] within Fairy's multi-layer planning architecture (Global Planner, Planner, Action Decider). We posit that the necessity and form of this interaction differ fundamentally across Fairy's abstraction levels:

\begin{itemize}
    \item \textbf{High-Level (Global Planner) Interaction:} We argue that at the Global Planner level (responsible for cross-app decisions), the runtime expectations defined by [RGR-II] (e.g., OR-decomposition or missing information) are rarely encountered. User intent ambiguity (e.g., "order a burger combo") is typically discovered within an application (mid/low levels), not during the high-level phase of "choosing which App." Therefore, the Global Planner's interaction is designed to be Trivial. Its primary duty is role allocation based on environmental knowledge. Its only interaction is an implicit failure report: if it cannot find a suitable app to fulfill the task decomposition, the system reports failure directly to the user, without requiring a specialized intent scaffold for clarification.
    
    \item \textbf{Mid/Low-Level (Planner \& Action Decider) Interaction:} In contrast, the mid-level (Planner, responsible for sub-goal planning) and low-level (Action Decider, responsible for atomic operations) are the primary venues for identifying runtime expectations. However, adhering to the software engineering DRY (Don't Repeat Yourself) principle, we did not build duplicate interaction modules for these two engines. Instead, we designed a unified, re-entrant user interaction module. This shared module serves both the mid and low levels, differing only in its activation (Triggers).
\end{itemize}

\begin{figure}[t]
  \centering
  \includegraphics[width=\linewidth]{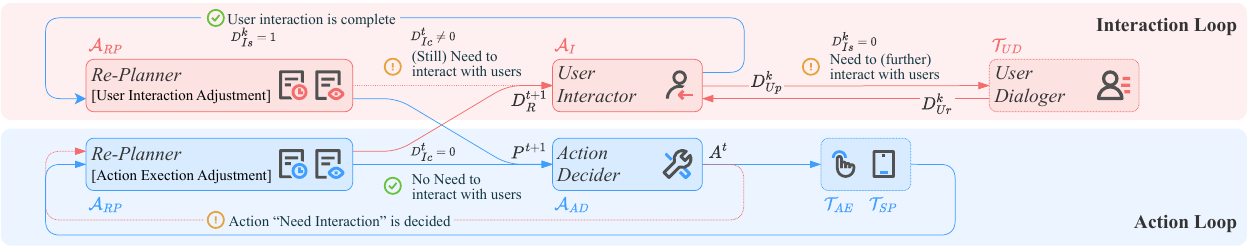}
  \caption{The primary workflow of the Interaction Loop. In the Interaction Loop, the User Interactor and Re-Planner agents collaborate with the user as needed, gathering feedback and adjusting the task execution strategy. The Interaction Loop also includes the User Dialoger tool, which provides a friendly interaction interface to ensure smooth communication with the user.}
  \label{fig:InteractionLoop}
\end{figure}

This design ensures that Fairy's architecture fully adheres to [RGR-II] while also guaranteeing high system maintainability from an engineering perspective. Corresponding to the Action Loop, this collaborative closed-loop——formed by runtime requirement discovery and User Interaction——is termed the Interaction Loop in the Fairy architecture, as shown in Figure \ref{fig:InteractionLoop}.

In the Action Loop, when $\mathcal{A}_{RP}$ determines that the user interaction type $D_{It}$ is not 0, the current Action Loop is suspended, and control is passed to the Interaction Loop for user dialogue. The possible values of $D_{It}$ are: 1- confirm a sensitive or potentially dangerous action; 2- confirm an irreversible action; 3- choose among multiple options; 4- clarify an instruction; 0- no interaction is required.

To prevent missed user interactions due to hallucinations from $\mathcal{A}_{RP}$, $\mathcal{A}_{AD}$ is allowed to select NeedInteraction as a correction. Once selected, $\mathcal{A}_{RP}$ can reflect on the error in the next loop and enter the Interaction Loop accordingly.

\noindent \textbf{The User Interactor ($\mathcal{A}_{I}$)} interprets the sub-goal from the $\mathcal{A}_{RP}$ to generate user prompts, receive replies, and determine whether the responses meet interaction needs. It summarizes the dialogue once all requirements are fulfilled. $\mathcal{A}_{I}$ is activated when the interaction type $D_{It}$ /0 or when a user reply $D_{R}$ is received. Given the task instruction $I_T$, current plan $P$, interaction request $D_{R}$, screen perception $S$, and previous context $C$, $\mathcal{A}_{I}$ generates the interaction plan $D$. If $k$ rounds of interaction have occurred, $\mathcal{A}_{I}$ additionally reads the user prompts $D_{Up}$ and responses $D_{Ur}$ from the interaction history $\{ \mathcal{M}_{I}^{\tau} \}_{\tau = 0}^{k}$.
$$D^{k+1}=\mathcal{A}_{I}({I}_{T},P^{t},D_{R}^{t},S^t,C^{t-1},\{ \mathcal{M}_{I}^{\tau} \}_{\tau = 0}^{k})$$
$D$ includes the user prompt $D_{Up}$ and interaction state $D_{Is}$. The possible values of $D_{Is}$ are: 0- further interaction is required; 1- the user has made a clear choice or clarification. When $D_{Is}$=1, a summary $D_{Us}$ is also generated by $\mathcal{A}_{I}$ for updating $I_{T}$.

\noindent\textbf{The User Dialoger ($\mathcal{T}_{UD}$)} is a tool for presenting user prompts $D_{Up}$ to the user $\mathcal{U}$ via a GUI interface or Console, and waiting for the user's response $D_{Ur}$. $\mathcal{T}_{UD}$ is activated when $D_{Is}$=0. During evaluation, $\mathcal{T}_{UD}$ can also be connected to an automated testing system.
$$D_{Ur} = \mathcal{T}_{UD}(D_{Up},\mathcal{U})$$
\noindent\textbf{The Planner ($\mathcal{A}_{RP}$)} is also responsible for reviewing the results of user interaction and deciding whether to revise or continue the current plan, in addition to the two stages introduced in the Action Loop:

\begin{enumerate}
    \setcounter{enumi}{2}
    \item \textbf{Adjustment (User Interaction):} $\mathcal{A}_{RP}$ examines whether the summarized user response $D_{Us}$ sufficiently covers the requested information in $D_{R}$ and the current screen perception $S$. Then, based on the reflection result, task instruction $I_T$, current plan $P$, and key context $C$, it determines whether further user interaction is needed and outputs a new request $D_R$, updates the $P_{Op}$, and selects the next $P_{Sg}$.
    $$P^{t+1}, D_{R}^{t+1}=\mathcal{A}_{RP}(D_{R}^{t},S^{t},D^t_{Us},{I}_{T},P^{t},C^{t-1})$$
    When the interaction type $D_{It}$, the Interaction Loop terminates and control returns to the Action Loop, where $\mathcal{A}_{AD}$ is invoked to decide the next atomic action.
\end{enumerate}

\subsection{Long- and Short-Term Memory Components of the EMA Specification}
\label{sec:ema_memory}

In Sections \ref{sec:oca_macro_architecture} and \ref{sec:rgr_planning_engines}, we instantiated the system's macro-level execution architecture and core planning engines by adhering to the OCA and RGR specifications. However, these components cannot yet operate collaboratively because the Task Knowledge, Environmental Knowledge, and Runtime Context required by [RGR-I] are not yet supplied. Therefore, in this section, we define the Memory System responsible for supplying and evolving this knowledge within Fairy's architecture.
EMA imposes two strict engineering requirements on this memory system: [EMA-I] mandates an explicit Layered Memory Structure, and [EMA-II] mandates an execute-evolve dual-loop model. It can be inferred that Fairy's memory system must contain the following components:

\begin{itemize}
    \item \textbf{Short-term Memory Manager}: A component defined by [EMA-I] for recording the Cognitive Stack during the Execution Loop and supporting high-frequency read/write access.
    \item \textbf{Long-term Memory Manager}: A component defined by [EMA-I] for persistently storing procedural and semantic knowledge, and providing a retrieval interface for Task Knowledge to the planning engines.
    \item \textbf{Evolution Loop}: A component defined by [EMA-II] responsible for post-hoc consolidation of experience from working memory into long-term memory, often implemented as a MAPE-K loop.
\end{itemize}

In the following sub-sections, we will demonstrate, layer by layer, how Fairy's various memory components implement the EMA specification.

\subsubsection{Short-term Memory and the Execution Loop.}
\label{sec:short_term_memory}

The Short-term Memory Manager is the core component for maintaining short-term memory. It manages all context relied upon and produced by Fairy's agents during task execution. Adhering to [EMA-I.1], the manager is designed as a composite stack structure, defined as follows:

\begin{itemize}
    \item \textbf{Instruction ($I_T$) Stack}: After the Task Manager assigns a task, the instruction $I_T$ is pushed onto the stack for use by agents within the App-Level Executor. When an Interaction Loop terminates, the user response summary $D_{Us}$ is also pushed, forming a supplemental update to $I_T$.
    \item \textbf{Action Loop Memory ($\mathcal{M}_{A}$) Stack}: When the $t$-th Action Loop completes, $$\mathcal{M}_{A}^t=\{P^t,A^t,S^t,R^t\}$$ is pushed onto the stack. A new memory $\mathcal{M}_{A}^{t+1}$ for the ($t+1$)-th round is then initialized at the top of the stack.
    \item \textbf{Interaction Loop Memory ($\mathcal{M}_{I}$) Stack}: When the $t$-th Action Loop is suspended, a new memory stack for that round, $\mathcal{M}_{I,t}$, is created. When the $k$-th Interaction Loop completes, $\mathcal{M}_{I,t}^k=\{D^k,D^k_{Ur}\}$ is pushed onto this stack. A new memory $\mathcal{M}_{I,t}^{k+1}$ is initialized at the top. After the Interaction Loop terminates, this stack $\mathcal{M}_{I,t}$ is cleared after its contents are cached.
    \item \textbf{Key Context ($C$) Stack}: When the $t$-th Action Loop concludes successfully and the Context Extractor has finished, $C^t$ is pushed onto the stack.
\end{itemize}

Globally, assume the App-Level Executor is handling the j-th sub-task. When $\mathcal{A}_{AD}$ outputs a Finish decision in the t-th Action Loop, the full execution record is stored as $$\mathcal{M}_{\mathcal{O}}^j=\{I_{T}^j,\{\mathcal{M}_A^{\tau},C^{\tau}\}_{\tau=0}^{t},\{\mathcal{M}_{I,\tau}^{(i)}\}_{\tau=0,\ i=0}^{t,\ n_\tau}\}$$ The key execution trace record is $\mathcal{M}_{T}^j=\{I_{T}^j,P^t_{Sg},\{A^{\tau},R^{\tau}\}_{\tau=0}^{t},C^t\}$. This key trace is used by the Global Planner ($\mathcal{A}_{GP}$) for task adjustment and by the Evolution Loop (Section \ref{sec:long_term_memory}).

Adhering to [EMA-I.3] and [EMA-II.1], the access pipelines for variables within these stacks are designed as follows:

\begin{figure}[t]
  \centering
  \includegraphics[width=\linewidth]{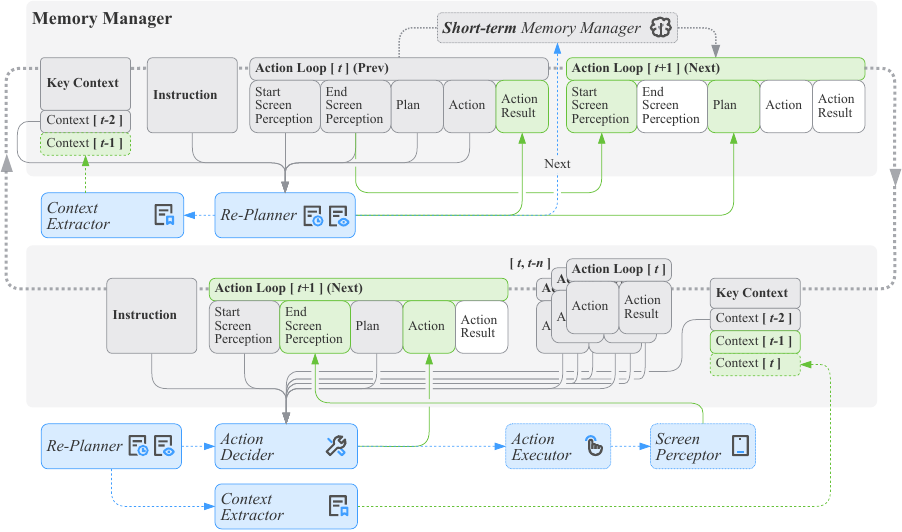}
  \caption{Memory scheduling logic among agents and tools in the Action Loop. The Re-Planner contributes the Action Result from the previous loop (t), then initializes the memory for the current loop (t+1) and adds the Plan. The Action Decider and Screen Perceptor contribute the Action and End Screen Perception for the current loop. Concurrently, the Context Extractor also contributes the Key Info t for this loop. As the loop progresses, new memory is continuously added, populated, and utilized.}
  \label{fig:Memeory-1}
\end{figure}

As shown in Figure \ref{fig:Memeory-1}, for the Action Loop, the Planner first retrieves from the memory manager the start screen, plan, action, and end screen of round $t$. It then generates the reflection result for round $t$ and produces the plan for round $t+1$. Once the reflection is completed, the memory for round $t$ is finalized. At this point, the memory manager initializes the memory for the next Action Loop round, using the end screen of round $t$ as the start screen of round $t+1$.
Next, the Action Decider receives the start screen and plan of round $t+1$, along with the execution traces from rounds $[t, t-n]$ (including only actions and their results), and determines the atomic action to perform. After the action is executed by the Action Executor, the Screen Perceptor is triggered to obtain the screen perception of round $t+1$. The Planner then repeats this process, advancing the loop to round $t+2$.
In parallel, when the Planner determines that round $t$ proceeded normally, the Context Extractor collects key information from the end screen of round $t$ (also the start screen of round $t+1$) during round $t+1$. This information is stored as the key context of round $t$. When executing round $t+1$, the system uses the key context from round $t-1$, because the context for round $t$ is still being collected. Moreover, since the start screen of round $t+1$ already includes information from round $t$, no redundant supplementation is needed.

\begin{figure}[t]
  \centering
  \includegraphics[width=\linewidth]{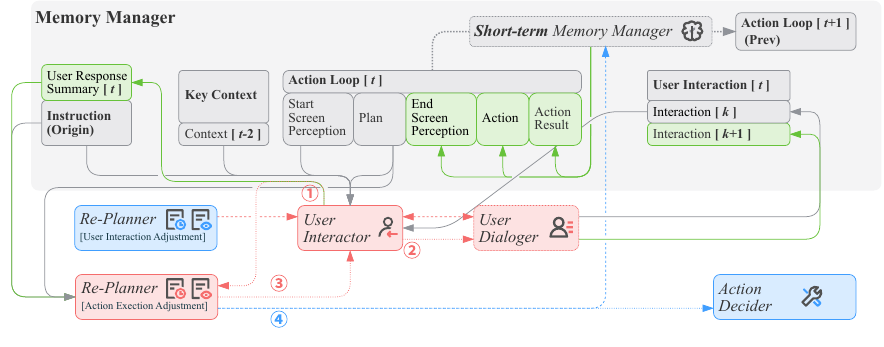}
  \caption{Memory scheduling logic among agents and tools in the Interaction Loop. If the Re-Planner discovers interaction is needed, the User Interactor will construct a user prompt and obtain the user's response via the User Dialoger. (1) If the user's reply does not meet the requirement, interaction continues, and the interaction is recorded. (2) Once the requirement is met, the User Interactor summarizes the user response and adds a User Updated Instruction; the Re-Planner will reflect on the interaction and, depending on the situation, either (3) adjust the goal to continue interaction or (4) initialize the memory for the next loop (t+1) and add the Plan. The Action Decider will complete the subsequent decision-making.}
  \label{fig: Memeory-2}
\end{figure}

As shown in Figure \ref{fig: Memeory-2}, during round $t$, if the Planner determines that user interaction is required, the Interaction Loop is triggered and engages in several rounds of communication with the user.
First, the User Interactor retrieves from the memory manager the start screen and plan of round $t$, along with user prompts and responses from previous interaction rounds [k, 1]. It then determines whether continued interaction is required. If the user's response still fails to meet the interaction requirement, the User Interactor generates the ($k+1$)-th prompt, which is displayed to the user via the User Dialoger. After receiving the response, the prompt and reply are stored together as the record of interaction round $k+1$, and the User Interactor continues planning for the next interaction round.
Once the user's response satisfies the requirement, the User Interactor summarizes the response and updates the Task Instruction accordingly. At this point, the Planner obtains the start screen, plan, and the updated Task Instruction of round $t$, reflects on the interaction, and generates a revised plan for round $t+1$ based on the newly clarified user requirements. After the reflection is complete, the memory manager retains empty values for any incomplete parts of the current Action Loop memory and initializes a new memory for round $t+1$, using the start screen from round $t$ as its start screen.
If the Planner still determines that user interaction is required, the Interaction Loop is reactivated. Otherwise, the Action Loop resumes, and the Action Decider begins selecting atomic actions.

\subsubsection{Long-term Memory and the Evolution Loop.}
\label{sec:long_term_memory}

The Long-term Memory Manager is the core component for maintaining long-term memory, managing the persistent knowledge that Fairy's agents rely upon. Adhering to [EMA-I.2], the manager is designed to store two categories of knowledge, defined as:

\begin{itemize}
    \item \textbf{App Tricks} are a series of experiential prompts for a certain App, belonging to the [EMA-I.2] defined procedural memory, including experience absorbed from previous executions, lessons absorbed from trial-and-error, and operation techniques provided by human experts. App Tricks are described by natural language text, including planning tricks ($T_{Plan}$), action decision tricks ($T_{Exec}$), and error recovery tricks ($T_{ExecErr}$).
    \item \textbf{App Map} is a knowledge graph for a certain App, belonging to the [EMA-I.2] defined Semantic Memory, including all App activities, pages, component information, and logical relationships seen in previous actions. App Map is described by a graph, each App includes several pages, which in turn include several in-page nodes, each page node includes the current component's description and operation logical relationship (e.g., jump to new page, partial refresh).
\end{itemize}

Adhering to [EMA-I.3] and [EMA-II.2], the query and update pipelines for these knowledge categories are designed as follows:

\noindent \textbf{1) App Tricks: } The App Trick Learner ($\mathcal{AL}_{AT}$) performs high-level reflection on the sub-task execution trace, identifying failed or redundant steps and discrepancies between initial and final plans. It summarizes experience-based tricks ($T$) to avoid similar issues and improve planning. After the App-Level Executor execution, $\mathcal{AL}_{AT}$ analyzes the full execution record $\mathcal{M}_{T}$ to generate three types of tricks for app $AM_p$: planning ($T_{\mathrm{p}}$), execution $T_{\mathrm{exe}}$, and error recovery ($T_{\mathrm{err}}$) .
$$\{\Delta T_{\mathrm{p}},\Delta T_{\mathrm{exe}},\Delta T_{\mathrm{err}}\}^{AM_p^j}= \mathcal{AL}_{AT}({I_{T}^j,P^t_{Sg},\{A^{\tau},S^{\tau},R^{\tau}\}_{\tau=0}^{t}})$$
Besides these learned tricks, $T$ also includes general-purpose tricks $T^\mathrm{Common}$ provided by human experts. At runtime, agents retrieve relevant knowledge via query $q$ using a retrieval-augmented generator (RAG, $\mathcal{S}$) :
$$T^{q,AM_p}_c=\mathcal{S}_c(T_c^{AM_p} \cup T_c^{\mathrm{Common}},q), c\in \{\mathrm{p},\mathrm{exe},\mathrm{err}\}$$

\begin{figure}[t]
  \centering
  \includegraphics[width=\linewidth]{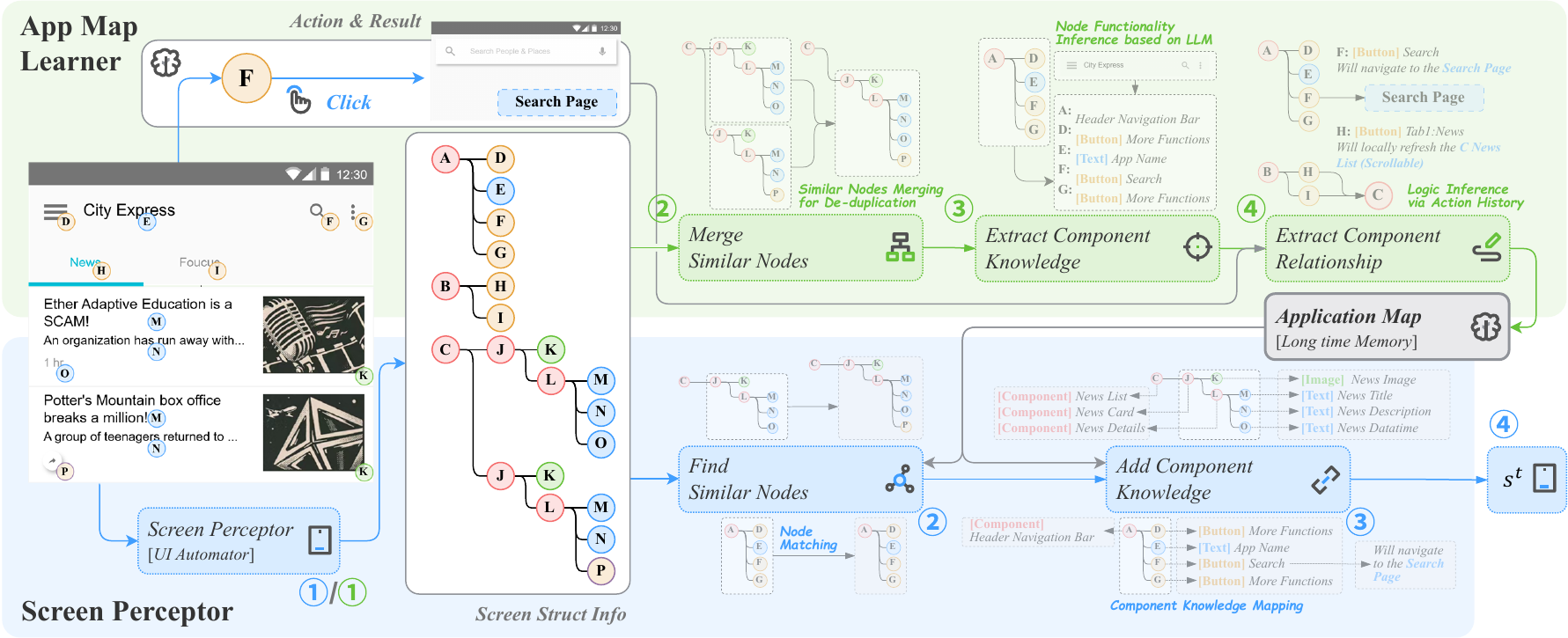}
  \caption{Workflow of the App-Map Learner. Green region -- knowledge learning: (l) perceive the UI structure, (2) matchand merge similar nodes, (3) extract component knowledge and (4) relationships; then update the long-term application map.Blue region - knowledge usage: (1) perceive the UI structure, (2) match the learned components; (3) iniect the component knowledge into the current UI structure, and (4) output it as screen perception $S^t$.}
  \label{fig:AppMapLearner}
\end{figure}

\noindent \textbf{2) App Map:} The App Map Learner ($\mathcal{AL}_{AM}$) tracks screen transitions, along with the triggering actions and components, to construct a knowledge graph of UI behavior. After execution, $\mathcal{AL}_{AM}$ analyzes each action execution trace $\mathcal{M}_{A}$ to generate descriptive records $M_\mathcal{N}$ for page components $\mathcal{N}$:
$$\{M_\mathcal{N}\}^{AM_p^j}= \mathcal{AL}_{AM}({S^{\tau-1},A^{\tau},S^{\tau}})$$

Specifically, the learning process of the $\mathcal{AL}_{AM}$ is illustrated in the green region of Figure \ref{fig:AppMapLearner}. After the App-Level Executor completes execution, $\mathcal{AL}_{AM}$ extracts screen perception data (e.g., screenshot and Accessibility Tree (AT)) from each action execution record $\mathcal{M}_{A}$. It then applies a tree similarity algorithm to match the current page against the app's page knowledge base. If no similar page is found, it is labeled as new; otherwise, $\mathcal{AL}_{AM}$ computes the differences and records new or modified component nodes (Step II). For each new page or component, $\mathcal{AL}_{AM}$ feeds the corresponding fragment and screenshot into an LMM to generate node descriptions (Step III). Then, based on the action logs in $\mathcal{M}_{A}$, it establishes links between UI changes and the components that triggered them (Step IV)(5). This information is saved (e.g., as a JSON file), detailing visited pages, component structures, functions, and trigger outcomes, providing concrete guidance to the Action Decider and avoiding reliance on commonsense, thereby improving its accuracy and stability.

Upon screen capture by $\mathcal{T}_{SP}$, matched components from $M$ are annotated into the textual screen perception $S_{Pt}$.

Specifically, the extraction process of the $\mathcal{AL}_{AM}$ is illustrated in the blue region of Figure \ref{fig:AppMapLearner}. After the Screen Perceptor obtains the current screen's AT, $\mathcal{AL}_{AM}$ applies the same tree similarity algorithm to match this screen with known pages in the app's page repository (step II in blue region). If a match is found, it retrieves the corresponding component descriptions and the outcomes of trigger actions (step III in blue region), and injects them into the current screen perception. In visual mode, this information is associated with the corresponding marks and output as additional textual perception $S_{Pt}$. In non-visual mode, the information is directly appended to the descriptions of the corresponding components in the output $S_{Pt}$.

\subsection{Micro-Implementation Architecture of the OCA Specification}
\label{sec:oca_micro_implementation}

In Sections \ref{sec:oca_macro_architecture} to \ref{sec:ema_memory}, we defined Fairy's planning engines and memory system. However, [OCA-II] imposes strict micro-engineering requirements on how these components collaborate. This aims to solve the black-box, tight coupling, and ad-hoc design problems prevalent in SoTA methods.
The core of this specification is cognitive decoupling and state-control separation. It has a direct implication for the system architecture: the system must manage \textbf{control flow} via an Event Bus (EB) and \textbf{data flow} via a Memory Bus (MB), while strictly prohibiting direct component-to-component calls.
To precisely instantiate [OCA-II], we designed and implemented a lightweight agent communication framework named Citlali. This framework is the cornerstone for achieving Fairy's white-box observability. Citlali's implementation strictly adheres to the implications of [OCA-II]:

\subsubsection{Cognitive Decoupling.}
    Citlali defines an abstract class Worker. In Fairy, all core cognitive components—including all Agents (e.g., Re-Planner, Action Decider), Tools (e.g., Action Executor, Screen Perceptor), and the Memory Manager—must be implemented as subclasses of Worker. This mandatorily decouples the macro-level agent into a series of single-responsibility, independently developable components.
    
\subsubsection{Event Bus and Memory Bus.}
   Citlali defines and implements two communication methods between components within the Worker abstract class: (1) Publish-Subscribe under a Topic, and (2) Request-Response by specifying a Worker's name. Fairy further instantiates this inter-component communication and coordination capability into the Event Bus and Memory Bus:
    
    \begin{itemize}
        \item \textbf{Event Bus (EB) / Control Flow:} All coordination between Fairy's Agents and Tools is based on the Publish-Subscribe model. For example, the Action Decider does not directly call the Action Executor. Instead, it publishes an atomic action to a specific topic, which the Action Executor subscribes to. This approach achieves the explicit control flow required by [OCA-II.2].
        
        \item \textbf{Memory Bus (MB) / Data Flow:} The Memory Manager is implemented as the \textbf{Memory Bus}. It listens (subscribes) to all content published by Agents and Tools via the Publish-Subscribe model and stores this information in short-term memory (as described in Section \ref{sec:short_term_memory}), thus serving as the single source of truth for the data flow. When an Agent or Tool needs historical state, it must acquire it from the Memory Manager via the Request-Response model. This achieves the explicit state flow required by [OCA-II.2].
    \end{itemize}
    
    This separation of Publishing Events (EB) and Querying State (MB) completely eradicates direct component-to-component calls and opportunistic polling, ensuring system debuggability and maintainability and achieving the deterministic coordination required by [OCA-II.3].

\section{Evaluation}
\label{ch:Evaluation}
Chapter \ref{ch:engineering_framework} introduced a systematic software engineering framework comprising Runtime Goal Refinement, Observable Cognitive Architecture , and Evolutionary Memory Architecture. This framework is designed to address the inherent non-determinism and adaptivity challenges posed by Agentic systems. In this chapter, we design a series of empirical experiments centered on Fairy, our prototype agent system implemented in Chapter \ref{ch:case_study_fairy}, which strictly adheres to the framework proposed in Chapter \ref{ch:engineering_framework}. Our objective is to quantitatively and qualitatively evaluate the effectiveness of this framework (and its constituent methodologies) in tackling the aforementioned challenges.
Our empirical evaluation is designed to answer the following core research questions:
\begin{itemize}
\item \textbf{(RQ-1):} Does the agent system constructed based on the three principles effectively improve task performance?
\item \textbf{(RQ-2):} Does adhering to the OCA macro-architecture and RGR specifications improve task performance?
\item \textbf{(RQ-3):} Does adhering to the EMA specification further improve task performance?
\item \textbf{(RQ-4):} Does adhering to the OCA micro-architecture specification improve system maintainability?
\end{itemize} 
\subsection{Evaluation Benchmarks\label{sec: evaluation_benchmarks}}
Given that Fairy is a mobile GUI agent assistant, our experimental design must evaluate its performance on relevant benchmarks to comprehensively answer our core RQs. This evaluation covers two aspects:
\begin{itemize}
\item \textbf{General Effectiveness:} To evaluate whether Fairy demonstrates fundamental task execution capabilities on standardized and widely accepted mobile agent benchmarks.
\item \textbf{Core Contribution Effectiveness:} To assess whether Fairy exhibits significant advantages on custom-designed benchmarks that specifically target its core engineering strengths, such as handling ambiguity, complex hierarchical tasks, and evolutionary learning.
\end{itemize}
To this end, we adopt a Dual-Benchmark strategy, combining a public dataset (AndroidWorld) with a custom-built dataset (RealMobile-Eval).

\subsubsection{General Benchmark - AndroidWorld.} In our experimental evaluation, we selected AndroidWorld as a general-purpose external benchmark. Maintained by Google, AndroidWorld provides a controllable simulation platform that closely aligns with real-world mobile environments, covering a set of typical mobile interaction tasks. It is widely used for evaluating the comprehensive performance of mobile agents. 
By conducting comparative experiments on AndroidWorld, we are able to validate the fundamental effectiveness and generalizability of our proposed design principles within a recognized and standardized environment.

\subsubsection{Challenging Benchmark - RealMobile-Eval.} The task scenarios in AndroidWorld primarily focus on static interface operations and single-turn task execution. This makes it difficult to fully evaluate the complete contributions of our framework, particularly its capabilities in handling highly complex and non-deterministic scenarios:

\begin{itemize}
\item \textbf{Lack of Complex Task Decomposition:} Tasks in AndroidWorld (e.g., productivity, communication) are largely confined within single applications. This restricts the evaluation of the OCA macro-architecture's advantages in processing complex, multi-stage, cross-application goals, consequently limiting our ability to adequately address RQ-3.
\item \textbf{Absence of Runtime Requirement Ambiguity:} Existing popular benchmarks, including AndroidWorld, suffer from a fundamental limitation: their tasks typically encapsulate all user requirements within the initial instruction. However, in realistic scenarios, users are unlikely to exhaustively specify all requirements at the outset (e.g., naming the specific app, desired store, and burger type all at once when ordering). This not only diminishes the task's realism but also prevents the core mechanisms of our RGR methodology—namely, runtime goal refinement, the differentiation between Runtime Requirements and runtime expectations, and the activation of User Interaction—from being triggered or evaluated. This severely constrains our ability to adequately address RQ-3.
\end{itemize}

To address the aforementioned limitations and to specifically evaluate the core value embodied by RGR and OCA, we propose \textbf{RealMobile-Eval}, a new, realistic, and challenging benchmark. This benchmark comprises 30 tasks designed by domain experts, based on real-world applications (e.g., Amazon, Zillow, McDonald's). These tasks feature intentionally ambiguous instructions and critical decision points, designed to test the agent's engineering robustness in complex, ambiguous, and dynamic scenarios.

\begin{table}[h!]
    \centering
    \caption{Classification of Task Difficulty }
    \label{tab:difficulty_criteria} 
    \begin{tabular}{l c c c L{5.5cm}}
        \toprule
        Difficulty Level & Single-App & Multi-App & Total & Criteria for Difficulty Classification \\
        \midrule 
        Simple  & 10 & 0 & 10 & Does not cross applications; User requirements are always clear; User requirements < 2; < 5 steps. \\
        Medium  & 10 & 4 & 14 & No cross-app or crosses 2 applications; Involves ambiguous user requirements; User requirements < 4; < 15 steps. \\
        Complex & 1  & 5 & 6  & Crosses more than 2 applications; Involves ambiguous user requirements; User requirements $\geq$ 4; $\geq$ 15 steps. \\
        \bottomrule 
    \end{tabular}
\end{table}

The core characteristics of tasks in RealMobile-Eval include:
\begin{itemize}
\item \textbf{Multi-stage, Cross-application Execution:} Medium and complex tasks are designed to require an extended sequence of operational steps and involve cross-application coordination, exhibiting a level of complexity that significantly surpasses previous benchmarks.
\item \textbf{ Clarification of Ambiguous Instructions:} Complex tasks are intentionally designed to contain ambiguous instructions and decision points (e.g., "Help me order a burger combo"). This design compels the system to proactively initiate interaction to seek clarification, rather than performing blind refinement.
\end{itemize}   

\subsection{Evaluation Methodology and Metrics}
\subsubsection{\textbf{Evaluation Methodology and Metrics for the AndroidWorld Benchmark}}

For the AndroidWorld benchmark, we first adapted its official evaluation paradigm. As discussed in Section \ref{sec: evaluation_benchmarks}, tasks in AndroidWorld primarily focus on validating fundamental application operation capabilities. To ensure a fair alignment with existing related work on core metrics and to validate the foundational effectiveness of our method, we fully adopted the primary metric advocated by this benchmark:
\begin{itemize}
    \item \textbf{Binary Success Rate($BSR$):} Measures whether a task is fully completed.
\end{itemize}
However, $BSR$ alone is insufficient for fine-grained failure analysis. Therefore, we additionally introduced the Step Redundant Ratio ($SRR$) as a supplementary diagnostic metric, which measures the proportion of redundant steps relative to the total steps within completed tasks.
\subsubsection{\textbf{RealMobile-Eval Evaluation Methodology: LLM-based Automated Evaluation and Interaction}}

When addressing the more complex RealMobile-Eval benchmark, we identified significant limitations in traditional evaluation metrics and methods:
\begin{itemize}
    \item \textbf{Limitations of Simple Metrics:} Much existing work relies on \textbf{Binary Success Rates} or matching against a \textbf{Ground Truth Trajectory}. However, real-world tasks (such as those in RealMobile-Eval) may have multiple valid completion paths and varying degrees of completion. Using these simplistic metrics fails to provide an accurate assessment.
    \item \textbf{Limitations of Manual Evaluation:} Although manual evaluation retains indispensable value in capturing complex human preferences, it can introduce subjective biases from evaluators, thereby compromising the reliability and validity of the research.
    \item \textbf{Lack of Interaction Capability Assessment:} There is currently a lack of evaluation methods specifically designed for interactive mobile assistants. Existing static benchmarks are incapable of assessing an agent's ability to proactively initiate interaction and seek clarification when faced with ambiguous requirements.
\end{itemize}
To address these limitations, we designed an automated evaluation pipeline based on LLMs (using GPT-4o\cite{OpenAI2024GPT4o} as the backbone model), which features a \textbf{separation of Evaluator and Driver roles}. This pipeline is specifically tailored for the RealMobile-Eval benchmark.
\begin{figure}[t]
  \centering
  \includegraphics[width=\linewidth]{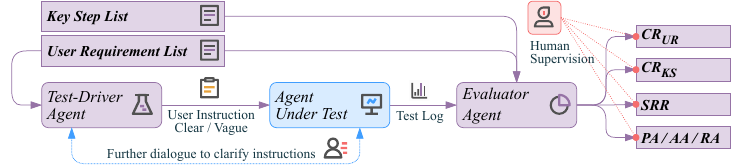}
  \caption{The Task Driver Agent is responsible for driving the test, constructing either a clear or an ambiguous instruction based on the User Requirement List, and continuously interacting with the agent; the Evaluator agent is responsible for analyzing the test logs to compute the relevant metrics.}
  \label{fig: test agent}
\end{figure}

    \noindent \textbf{Task Driver Agent:} Simulates a real user by issuing instructions to the system under evaluation and responding to its interaction requests.
    \begin{itemize}
        \item \textbf{Evaluation Process:} For each task, we compiled a \textbf{User Requirement List} and a \textbf{Key Step List}. The Task Driver Agent then outputs either a clear or an ambiguous user instruction (based on the lists and task difficulty) for the evaluated system to execute.
        \item \textbf{Interaction Support:} For ambiguous requirements, when the evaluated system (e.g., Fairy) requests user interaction, the Task Driver Agent reads the provided user prompt (such as RGR's intent scaffolding) and provides a clarifying response based on the User Requirement List. This mode is not activated when the interactive agent faces clear requirements. This process is supervised by human experts to ensure the response contains no additional content (e.g., extra operational step information) that could influence task completion, beyond supplementing the requirements.
    \end{itemize}
    
    \noindent \textbf{Task Evaluation Agent:} Acts as the Evaluator to automatically score the performance of the evaluated system after task execution.
    \begin{itemize}
        \item \textbf{Evaluation Process:} For each task, the Task Evaluation Agent assesses the number of completed user requirements and key steps based on the User Requirement List, the Key Step List, and the full task execution records (including screen captures and operation logs).
        \item \textbf{Reliability Assurance:} To mitigate the impact of LLM hallucinations, the Task Evaluation Agent is required to provide specific justifications for all its judgments. To ensure the Task Evaluation Agent was designed correctly, prior to the formal evaluation, we assembled a panel of five experienced mobile application testers. We randomly sampled 15\% of the test results and invited human experts and the Task Evaluation Agent to assess them concurrently. We used Inter-Rater Reliability (IRR) to ensure consistency in definitions, and the full evaluation commenced only after the Kappa coefficient exceeded 0.85. Additionally, all evaluation results are subject to secondary screening by human experts from the group, and when an incorrect judgment by the LLM is identified, human experts intervene to correct it.
    \end{itemize}
\subsubsection{\textbf{Evaluation Metrics for the RealMobile-Eval Benchmark}}

After task execution, the Task Evaluation Agent also computes the following evaluation metrics (this computation does not rely on the model's capabilities):
\begin{itemize}
    \item \textbf{User Requirement Completion Rate ($CR_{UR}$) :} The proportion of completed user requirements to the total user requirements.
    \item \textbf{Key Step Completion Rate ($CR_{KS}$) :} The proportion of completed key steps to the total key steps.
    \item \textbf{Step Redundancy Rate ($SRR$) :} The proportion of redundant steps to the total steps within completed tasks.
\end{itemize}
Furthermore, specifically for our work (the Fairy system), the Task Evaluation Agent additionally computes the following \textbf{Diagnostic Metrics} to evaluate the internal performance of the key cognitive components within our RGR, OCA, and EMA framework:
\begin{itemize}
    \item \textbf{Planning Error Rate ($PER$) :} The proportion of steps with planning errors to the total steps.
    \item \textbf{Action Error Rate ($AER$) :} The proportion of steps with action errors to the total steps.
    \item \textbf{Reflection Error Rate ($RER$) :} The proportion of steps with reflection errors to the total steps.
\end{itemize}
Based on the methodology described above, we ensure \textbf{comparability} for the evaluation on AndroidWorld, as well as the \textbf{ecological validity} and \textbf{fine-grainedness} of the evaluation on RealMobile-Eval.

\subsection{Experimental Setup and Evaluation Plan }\label{sec 5.3}

\subsubsection{Systems Under Evaluation.}\label{sec 5.3.1}

\begin{enumerate}
    \item \textbf{SoTA Baselines}: We first conducted a systematic comparison against state-of-the-art (SoTA) mobile agent frameworks. We selected several of the most popular open-source mobile agent systems as baselines, including App-Agent \cite{Li2024AppAgentV2}, Mobile-Agent-V2 \cite{Wang2024MobileAgentV2}, Mobile-Agent-E \cite{Wang2025MobileAgentE}, MobA \cite{Zhu2025MobA},and M3A\cite{Rawles2024AndroidWorld}.
    
    To precisely evaluate the existing implementations of these SoTA frameworks, we utilized the six core pillars articulated in Chapter \ref{ch:engineering_framework} as a retrospective analytical lens to deconstruct the core mechanisms that have independently evolved within these baselines. We adhered to the principle of fair alignment: if a baseline system possesses a feature—independently evolved based on existing AI principles—that aligns in core ideology with the engineering requirements of one of our pillars, we recognize it as having the corresponding capability in that dimension. We do not require the baseline to have implemented the entire specification of that pillar. Table\ref{tab:agent_comparison} reveals the architectural implementation differences between these methods and Fairy.
   
    \item \textbf{Fairy:} The mobile agent system proposed in Chapter \ref{ch:case_study_fairy}, which systematically and completely implements all six engineering pillars.
        
    \item \textbf{Fairy (Non-EMA):} This is an \textbf{ablated version} of Fairy. It fully implements the RGR and OCA architectures but removes the long-term memory and the EMA-II evolutionary execution dual-loop from the EMA framework. This baseline is specifically used in RQ4 to \textbf{isolate and evaluate} the contribution of the EMA specification.
\end{enumerate}
\begin{table}[htbp]
    \centering 
    
    \renewcommand{\arraystretch}{1.1} 
    \setlength{\abovetopsep}{0pt}
    \setlength{\belowrulesep}{0pt}
    \setlength{\aboverulesep}{0pt}
    \setlength{\belowbottomsep}{0pt}
    
    \caption{Comparison of Agent Architectures}
    \label{tab:agent_comparison}

    \begin{tabular}{>{\scriptsize}l S{3.75cm} S{3.75cm} S{3.75cm}}
        \toprule
        
        \textbf{Feature} & \textbf{M3A} & \textbf{App-Agent} & \textbf{Mobile-Agent-V2} \\
        \midrule
        
        RGR-I & $\times$~Partial (Relies on runtime context \& environmental knowledge) & \checkmark~(Relies on AppDoc, runtime context \& environmental knowledge) & $\times$~Partial (Relies on runtime context \& environmental knowledge) \\
        RGR-II & $\times$~(Single-turn interaction) & $\times$~(Single-turn interaction) & $\times$~(Single-turn interaction) \\
        OCA-I & $\times$~(Monolithic architecture) & $\times$~(Monolithic architecture) & \checkmark~(2-layer multi-agent architecture) \\
        OCA-II & $\times$~(Monolithic architecture) & $\times$~(Monolithic architecture) & $\times$~(Tightly-coupled multi-agent) \\
        EMA-I & $\times$~(Short-term memory only) & \checkmark~(Long-Short-term memory hierarchy) & $\times$~(Short-term memory only) \\
        EMA-II & $\times$~(No evolutionary loop) & $\times$~Partial (Offline-phase learning) & $\times$~(No evolutionary loop) \\
        
        \cmidrule{2-4}
        
        \textbf{ } & \textbf{Mobile-Agent-E} & \textbf{MobA} & \textbf{Fairy}  \\
        
        \cmidrule{2-4}
        
        RGR-I & \checkmark~(Relies on Tips/Shortcuts, runtime context) & \checkmark~(Relies on historical memory, runtime context) & \checkmark~(Relies on Tricks/AppMap, runtime context)  \\
        RGR-II & $\times$~(Single-turn interaction) & $\times$~(Single-turn interaction) & \checkmark~(Full Interaction capability)  \\
        OCA-I & \checkmark~(2-layer multi-agent architecture) & \checkmark~(2-layer multi-agent architecture) & \checkmark~(3-layer multi-agent architecture)  \\
        OCA-II & $\times$~(Tightly-coupled multi-agent) & $\times$~(Tightly-coupled multi-agent) & \checkmark~(Loosely-coupled via Publish-Subscribe)  \\
        EMA-I & \checkmark~(Long-Short-term memory hierarchy) & \checkmark~(Long-Short-term memory hierarchy) & \checkmark~(Long-Short-term memory hierarchy)  \\
        EMA-II & \checkmark~(Includes self-learning component) & $\times$~Partial (Historical memory replay) & \checkmark~(Includes self-learning component)  \\
        
        \bottomrule
    \end{tabular}
\end{table}

\subsubsection{RQ Evaluation Plan.\label{sec: 5.3.2}}

Based on the systems under evaluation defined in Section \ref{sec 5.3.1}(SoTA Baselines, Fairy, Fairy (Non-EMA)), we designed the following experimental plans, intended to answer our core research questions individually.
    
\noindent \textbf{1) RQ1 Evaluation Plan:} For RQ1, we conduct a comprehensive evaluation of the aforementioned methods on both AndroidWorld (full set) and RealMobile-Eval (full set):

\begin{itemize}
    \item \textbf{On the AndroidWorld Dataset:} As a general external benchmark, we only evaluate Fairy on this dataset to provide comparability with other related works.
    
    \item \textbf{On the RealMobile-Eval Dataset:} As our new challenging benchmark, we evaluate all systems on this dataset. Since all baseline methods (except Fairy) support only \textbf{single-turn interaction}, the Task Driver Agent provides them with clear user instructions that encapsulate all requirements at once. Conversely, Fairy is subjected to a more challenging evaluation: the Task Driver Agent first issues ambiguous user instruction, and then provides further requirements progressively based on Fairy's interactions. All methods are evaluated on Simple, Medium, and Complex tasks.

\end{itemize}
        
\noindent \textbf{2) RQ2 Evaluation Plan:} For RQ2, we decompose it into the following three sub-questions for further investigation:

\begin{itemize}
    \item RQ2-1: Does the OCA macro-architecture improve the user requirement completion rate?
    \begin{itemize}
        \item \textbf{Evaluation Plan:} For RQ2-1, we select App-Agent, Mobile-Agent-E, and Fairy, conducting a comparative analysis on the Medium and Complex tasks from RealMobile-Eval under Precise User Instructions (to preclude the effects of the RGR-II specification).
                
        \item \textbf{Fairness Analysis:} As shown in Table \ref{tab:agent_comparison}, in this setup, these systems only differ in the number of agent layers (1-layer, 2-layer, and 3-layer, respectively; the OCA-I specification does not affect this result), thus constituting a fair evaluation.
    \end{itemize}

    \item RQ2-2: Does the RGR-I specification effectively improve task performance?
    \begin{itemize}
        \item \textbf{Evaluation Plan:} For RQ2-2, we select Mobile-Agent-V2, Mobile-Agent-E, and Fairy, conducting a comparative analysis on the Simple tasks from RealMobile-Eval under Precise User Instructions (to preclude the effects of the RGR-II specification).
                
        \item \textbf{Fairness Analysis:} As shown in Table \ref{tab:agent_comparison} , all selected methods rely on contextual knowledge. Mobile-Agent-V2 relies only on the CV-based perception method FVP, not task decomposition knowledge. Mobile-Agent-E relies on task decomposition knowledge and the same FVP. Fairy also relies on task decomposition knowledge but uses the UI Tree-based SoM perception method SSIP. In this test, by selecting Simple tasks, Fairy's global planner (responsible for cross-application tasks) only performs trivial operations. Therefore, these systems exhibit no difference in agent layers and differ only in the knowledge used for task decomposition (the OCA-I specification does not affect this result), thus constituting a fair evaluation.
    \end{itemize}
    
    \item RQ2-3: Does the RGR-II specification effectively improve the user requirement completion rate?
    \begin{itemize}
        \item \textbf{Evaluation Plan:} For RQ2-3, we select App-Agent, Mobile-Agent-V2, Mobile-Agent-E, MobA, and Fairy, conducting a comparative analysis on the Medium and Complex tasks from RealMobile-Eval under ambiguous user instructions.
                
        \item \textbf{Fairness Analysis:} As shown in Table \ref{tab:agent_comparison}, all SoTA baselines are limited to single-turn interaction, whereas only Fairy possesses runtime interaction capabilities. A comprehensive comparison is necessary to reveal the critical role of the RGR-II specification when facing ambiguous user instructions in realistic scenarios.
    \end{itemize}
\end{itemize}

For all sub-questions above, if the relevant evaluation results are already contained within the experiments for RQ-1, no additional experiments are conducted.
        
\noindent \textbf{3) RQ3 Evaluation Plan:} For RQ3, we conduct an ablation study of Fairy on both AndroidWorld and RealMobile-Eval:
        
We select Fairy and its ablated version, Fairy (Non-EMA). We use AndroidWorld to assess their capability differences on single-application tasks, and RealMobile-Eval (Medium and Complex tasks) to assess their differences on multi-application tasks. For this evaluation, Fairy is required to perform three repeated executions of the tasks to accumulate knowledge; only the final task execution is used for the comparative evaluation.
        
\noindent \textbf{4) RQ4 Evaluation Plan:} Unlike RQ1, RQ2, and RQ3, RQ4 addresses a software engineering quality concern. We opt to evaluate this via a Human-Subject Study:

\begin{itemize}
    \item \textbf{Evaluation Plan:} To strictly control for the confounding effects of individual variance on the experimental results, we employ a Within-Subjects Design supplemented with Counterbalancing. We invited 6 experienced agent developers. Each expert was tasked with adding a new module—functionally similar to Fairy's user interaction module—to two architecturally divergent SoTA baselines (App-Agent and Mobile-Agent-E). The experts were randomly assigned to two groups: Group A performed the tasks in the order of App-Agent -> Mobile-Agent-E; Group B followed the reverse order. Concurrently, the time required to develop this module in Fairy was used as a comparative benchmark. All experts were instructed not to use AI-assisted development tools.
    
    \item \textbf{Evaluation Metrics:} RQ4 utilizes distinct evaluation metrics. We record and compare the average task completion time (H) required by the experts on the baseline systems (i.e., App-Agent and Mobile-Agent-E). Furthermore, to deeply diagnose the specific advantages of the OCA-I specification, this total time is further decomposed into three sub-stage durations to assess its improvement on software engineering practices:
            
    \textbf{Code Comprehension Time:} Assesses the impact of the system's architectural style on the time required to understand the code.
                
    \textbf{Agent Development Time:} Assesses the impact of the system's Cognitive Decoupling on development time.
                
    \textbf{Agent Debugging Time:} Assesses the impact of the system's Observability on error remediation and debugging time
\end{itemize}

\subsubsection{Evaluation Environment.}
All our experiments were run in an Android 15 environment. Apart from model configurations, all baseline methods followed their recommended settings. For Fairy's configuration, please refer to Table \ref{tab:my_config}.
\begin{table}[h!]
    \centering
    \caption{Experimental Configuration Settings }
    \label{tab:my_config}
    \begin{tabular}{ll}
        \toprule
        \textbf{Configuration Item} & \textbf{Value} \\
        \midrule
        Action Executor Type & UI AUTOMATOR \\
        Screenshot Getter Type & UI AUTOMATOR \\
        Screen Perception Type & SSIP \\
        Interaction Mode & CONSOLE \\
        Non-visual Execution Mode & False \\
        Manual Application Info Collection & True \\
        Reflection Policy & Hybrid \\
        \bottomrule
    \end{tabular}
\end{table}

To ensure all experiments (including SoTA baseline comparisons and ablation studies) are conducted on a fair basis, all systems under evaluation utilize a uniform configuration for their backbone and auxiliary models (where applicable).
The specific configuration is as follows:
\begin{itemize}
    \item \textbf{Backbone Model:} The system's core cognitive and planning capabilities are provided by \textbf{chatgpt-4o-2024-11-20}.
    \item \textbf{RAG and Embedding Models:} The Retrieval-Augmented Generation (RAG) model uses \textbf{qwen-turbo-0428}\cite{Alibaba2025QwenTurbo0428}; the Retrieval Embedding Model utilizes \textbf{intfloat/multilingual-e5-large-instruct}\cite{Wang2024MultilingualE5}.
    \item \textbf{Vision and Perception Models:} The Visual Prompt Model employs \textbf{qwen-vl-plus}\cite{Alibaba2025QwenVLPlus}.
\end{itemize}
All other models not specified herein followed their recommended configurations.

\section{Result \& Discussion}
\label{ch:Result&Discussion}

\subsection{RQ1: Does the Multi-Agent System Constructed Based on the Three Principles Effectively Improve Task Performance?}

To answer RQ1, we first evaluate the overall performance of Fairy as an integrated system incorporating the RGR, OCA, and EMA methodologies, comparing it against SoTA baselines. This RQ aims to establish a high-level baseline to validate the holistic effectiveness of the integrated principles. The subsequent RQs (RQ2, RQ3, and RQ4) will then delineate the specific attributions for this success.
We employed a Dual-Benchmark strategy, though the evaluation objectives and the systems under evaluation differed for each benchmark:
We first validated Fairy's foundational effectiveness on AndroidWorld, a public, standardized mobile agent benchmark. On this benchmark, we only compared Fairy against M3A (the baseline agent provided by the AndroidWorld framework). We did not evaluate all SoTA baselines (e.g., App-Agent, MobA) on this benchmark, primarily for two reasons:

\textbf{1) Mismatch in Evaluation Objectives}: As argued in Section \ref{sec: evaluation_benchmarks}, the AndroidWorld task set lacks the complexity necessary to evaluate our core contributions.

\textbf{2) Cost-Benefit Trade-off}: Porting and adapting all SoTA baselines to this public benchmark (which has low relevance to our research objectives) represents a high-cost, low-value engineering investment. Therefore, we opted to compare only against the benchmark's native baseline (M3A) to demonstrate that Fairy possesses solid foundational capabilities.

To genuinely evaluate our framework, we constructed and open-sourced our own RealMobile-Eval benchmark. As this benchmark was specifically designed by us to measure hierarchical complexity and interactive ambiguity, we invested the necessary engineering effort to adapt all relevant SoTA baselines (App-Agent, Mobile-Agent-V2, Mobile-Agent-E, MobA) to it for a comprehensive performance comparison. As described in Section \ref{sec: 5.3.2}, we employed an asymmetric evaluation on RealMobile-Eval: all SoTA baselines received clear instructions, while Fairy received the more difficult ambiguous instructions and progressively sought clarification from the Task Driver Agent.

Fairy's performance on the general benchmark (AndroidWorld) is presented in Table \ref{tab:androidworld}. Compared with M3A, Fairy demonstrated a significant improvement in binary success rate (\textit{BSR} increased by 111.8\%). This confirms that Fairy possesses solid foundational capabilities on a standard benchmark.

As shown in Table \ref{tab:realmobile-eval}, on the more challenging RealMobile-Eval benchmark, Fairy exhibited comprehensive performance advantages across all metrics. Compared to the best-performing SoTA baseline (Mobile-Agent-E), the User Requirement Completion Rate ($CR_{UR}$) improved by 33.7\%, and the Key Step Completion Rate ($CR_{KS}$) improved by 27.2\%. It must be emphasized that Fairy achieved this advantage under more stringent testing conditions; its $CR_{UR}$ under ambiguous requirements still surpassed all baselines operating under clear requirements.This robustly demonstrates the resilience of the integrated three-principle framework.

Finally, to verify that the three-principle framework is independent of any specific LMM, we tested Fairy on RealMobile-Eval using several different backbone LMMs. As shown in Table \ref{tab:model-comparison}, when using various LMMs, Fairy's internal cognitive components (Planning, Action, Reflection) all maintained low error rates (PER, AER, and RER all below 29.6\%). This indicates that the engineering specifications of our framework (RGR, OCA, EMA) provide structural advantages that are not dependent on a specific model.

\begin{table}[h!]
    \centering
    \caption{AndroidWorld Experiment Results}
    \label{tab:androidworld}
    \begin{tabular}{l cc}
        \toprule
        \multirow{2}{*}{Agent} & \multicolumn{2}{c}{AndroidWorld} \\
        \cmidrule(lr){2-3}
        & BSR & SRR \\
        \midrule
        M3A   & 30.6\% & - \\
        Fairy & 64.8\% & 14.7\% \\
        \bottomrule
    \end{tabular}
\end{table}

\begin{table}[h!]
    \centering
    \caption{RealMobile-Eval Experiment Results}
    \label{tab:realmobile-eval}
    \begin{tabular}{l ccc ccc ccc}
        \toprule
        \multirow{3}{*}{Agent} & 
        \multicolumn{9}{c}{RealMobile-Eval} \\
        \cmidrule(lr){2-10}
        & \multicolumn{3}{c}{Simple} & \multicolumn{3}{c}{Medium} & \multicolumn{3}{c}{Complex} \\
        \cmidrule(lr){2-4} \cmidrule(lr){5-7} \cmidrule(lr){8-10}
        & $CR_{UR}$ & $CR_{KS}$ & $SRR$ & $CR_{UR}$ & $CR_{KS}$ & $SRR$ & $CR_{UR}$ & $CR_{KS}$ & $SRR$ \\
        \midrule
        App-Agent         & 36.1\% & 63.1\% & 20.3\% & 37.8\% & 49.1\% & 15.4\% & 18.1\% & 19.4\% & 33.9\% \\
        Mobile-Agent-V2   & 68.2\% & 74.6\% & 30.2\% & 42.9\% & 49.8\% & 43.2\% & 29.8\% & 36.1\% & 42.0\% \\
        Mobile-Agent-E    & 72.7\% & 84.1\% & 22.3\% & 66.6\% & 71.5\% & 22.3\% & 38.2\% & 43.8\% & 55.7\% \\
        MobA              & 50.0\% & 62.2\% & 12.4\% & 46.4\% & 46.1\% & 24.2\% & 26.4\% & 27.1\% & 22.0\% \\
        Fairy             & 95.5\% & 98.5\% &  1.5\% & 83.3\% & 89.7\% & 17.7\% & 67.9\% & 75.6\% & 20.9\% \\
        \bottomrule
    \end{tabular}
\end{table}

\begin{table}[h!]
    \centering
    \caption{RealMobile-Eval Experiment Results For Different Models}
    \label{tab:model-comparison}
    \begin{tabular}{l c c c c c c}
        \toprule
        Model & $CR_{UR}$ & $CR_{KS}$ & $SRR$ & $PER$ & $AER$ & $RER$ \\
        \midrule
        ChatGPT-4o & 90.0\% & 93.2\% & 13.1\% & 7.8\% & 9.3\% & 7.8\% \\
        DeepSeekV3 & 83.3\% & 84.0\% & 15.1\% & 10.0\% & 12.4\% & 10.9\% \\
        DeepSeekR1 & 67.5\% & 71.6\% & 21.0\% & 21.9\% & 29.6\% & 18.0\% \\
        Qwen3      & 76.7\% & 79.7\% & 18.0\% & 12.5\% & 17.3\% & 15.2\% \\
        \bottomrule
    \end{tabular}
\end{table}

\begin{conclusion}{RQ1}
The consolidated findings from RQ1 confirm the holistic effectiveness of the multi-agent system built upon the three principles, demonstrating its capabilities on both general-purpose benchmarks and in complex real-world tasks. However, this aggregate success does not illuminate the internal mechanisms at play. We will, therefore, proceed with an in-depth analysis in the subsequent RQ2, RQ3, and RQ4 to dissect, isolate, and validate the independent contributions of OCA, RGR, and EMA, respectively.
\end{conclusion}

\subsection{RQ2: Does Adhering to the OCA and RGR Improve System Maintainability?}

We primarily focus on the $CR_{UR}$ (User Requirement Completion Rate) and $CR_{KS}$ (Key Step Completion Rate) metrics, as they measure the system's ability to satisfy top-level user requirements and complete critical sub-steps, respectively.

\subsubsection{\textbf{RQ2-1: Does the OCA Improve Task Performance?}}

\begin{table}[h!]
    \centering
    \small
    \caption{Performance Of Different Agent Architectures In Medium And Complex RealMobile-Eval Experiment}
    \label{tab:key-diff}
    \begin{tabular}{l c c c c L{6cm}}
        \toprule
        \multirow{2}{*}{Agent} & 
        \multicolumn{2}{c}{Medium} &
        \multicolumn{2}{c}{Complex} &
        \multirow{2}{*}{Key Difference} \\
        \cmidrule(lr){2-3} \cmidrule(lr){4-5}
        & $CR_{UR}$ & $CR_{KS}$ & $CR_{UR}$ & $CR_{KS}$ & \\
        \midrule
        App-Agent &
        37.8\% & 49.1\% &
        18.1\% & 19.4\% &
        Monolithic decision. Decomposes user instructions into atomic operations inside a single agent. \\ \cmidrule(lr){6-6}

        Mobile-Agent-E &
        66.6\% & 71.5\% &
        38.2\% & 43.8\% &
        Two-layer decision. Manager decomposes instructions into sub-goals. Operator decomposes sub-goals into atomic operations. \\ \cmidrule(lr){6-6}

        Fairy &
        83.3\% & 89.7\% &
        67.9\% & 75.6\% &
        Three-layer decision. Global Planner makes sub-tasks across apps. Planner makes sub-goals. Action Decider makes atomic operations. \\
        \bottomrule
    \end{tabular}
\end{table}

To answer RQ2-1, we evaluated the impact of the OCA-I on task performance. We selected App-Agent (monolithic architecture), Mobile-Agent-E (two-layer architecture), and Fairy (three-layer architecture) and compared their performance on the Medium and Complex tasks within the RealMobile-Eval dataset.

We selected the Medium and Complex tasks from RealMobile-Eval because they were specifically designed to include multi-stage, cross-application coordination. They serve as effective test cases for evaluating macro-architectural capabilities, such as logical stratification and cognitive delegation. In contrast, the Simple tasks (from both AndroidWorld and RealMobile-Eval) only involve single applications and thus fail to trigger Fairy's high-level planning mechanisms (like the Global Planner); they were therefore excluded from this comparison.Summary Table \ref{tab:key-diff} shows the comparison results, which clearly demonstrate that the logical stratification of the RGR planning engine has a decisive impact on the agent's ability to handle complex tasks.

\textbf{1) Monolithic Architecture (App-Agent) :} This architecture performed the worst. It attempts to decompose high-level user instructions into atomic operations in a single pass. This validates our hypothesis that a single planning engine, when faced with tasks requiring cross-application coordination, cannot effectively manage and apply knowledge at different levels of abstraction, leading to planning failure at its root.

\textbf{2) Two-Layer Architecture (Mobile-Agent-E) :} This architecture showed an improvement over the monolithic one ($CR_{UR}$ improved by 82.2\%; $CR_{KS}$ by 57.1\%). Although its Manager-Operator model is theoretically superior to a monolithic design, its high-level Manager still needs to manage non-specialized task knowledge. This indicates that a simple two-layer division is still insufficient to handle real-world complexity, causing it to fail frequently in multi-stage tasks.

\textbf{3) Three-Layer Architecture (Fairy) :} Fairy significantly outperformed the other two, showing further improvement over the two-layer architecture ($CR_{UR}$ improved by 35.5\%; $CR_{KS}$ by 35.3\%). This significant advantage stems from the specialization of responsibilities enforced by the OCA-I. Fairy strictly follows OCA-I, implementing cognitive delegation through three logically stratified RGR engines: The Global Planner (responsible for cross-app decisions), the Planner (responsible for intra-application functional orchestration), and the ActionDecider (responsible for atomic operation planning). This allows each of Fairy's RGR engines to precisely focus on knowledge at its specific level of abstraction, thereby making more accurate decisions.

\begin{conclusion}{RQ2-1}
Adhering to the OCA-I significantly improves an agent's performance on complex, multi-stage, cross-application tasks through effective task decomposition and reliable hierarchical collaboration.
\end{conclusion}

\subsubsection{\textbf{RQ2-2: Does the RGR-I Improve Task Performance?}}
\begin{table}[h!]
    \centering
    \small
    \caption{The impact Of RGR-I On Agent Capabilities}
    \label{tab:key-diff-RGR-I}
    \begin{tabular}{l c c c L{3.3cm} L{3.3cm}} 
        \toprule
        \multirow{3}{*}{Agent} & 
        \multicolumn{3}{c}{RealMobile-Eval} &
        \multicolumn{2}{c}{Key differences} \\
        \cmidrule(lr){2-4} \cmidrule(lr){5-6}
        &  \multicolumn{3}{c}{Simple} & 
        \multirow{2}{=}{RGR-I.2} & 
        \multirow{2}{=}{RGR-I.3}\\
        \cmidrule(lr){2-4}
        & $CR_{UR}$ & $CR_{KS}$ & $SSR$ &  & \\ 
        \midrule
        Mobile-Agent-V2 &
        68.2\% & 74.6\% & 30.2\% &
        Relying on screen perception info from FVP: \newline
        1) All textual info and icon descriptions identified via computer vision, along with their screen coordinates; \newline
        2) Screenshots annotated with point markers. &
        Does not rely on task decomposition knowledge: \newline
        lacks self-evolution mechanism \\ 
        \cmidrule(lr){5-5}\cmidrule(lr){6-6}

        Mobile-Agent-E &
        72.7\% & 84.1\% & 22.3\% &
        Relying on screen perception info from FVP, consistent with Mobile-Agent-E. &
        Relying on task decomposition knowledge: AppTips from self-evolution mechanism.\\ 
        \cmidrule(lr){5-5}\cmidrule(lr){6-6}

        Fairy &
        95.5\% & 98.5\% & 1.5\% &
        Relying on screen perception info from SSIP: Screenshots marked with SoM label boxes, based on UI Tree-identified elements. &
        Relying on task decomposition knowledge: AppTricks from self-evolution mechanism.\\
        \bottomrule
    \end{tabular}
\end{table}

To investigate RQ2-2, we evaluate the impact of RGR-I on task performance. We select Mobile-Agent-V2, Mobile-Agent-E, and Fairy, and compare their performance on the Simple Tasks subset of RealMobile-Eval. As described in RQ2-1, when executing Simple Tasks, Fairy’s three-layer architecture automatically degenerates into a two-layer architecture (i.e., the Global Planner only intervenes trivially), making its macro-architecture aligned with the baselines.
In this way, we eliminate the confounding effects of the OCA-I. As summarized in Table \ref{tab:key-diff-RGR-I}, under controlled conditions, the key differences among the three lie solely in how they implement RGR-I.2 and RGR-I.3 within the RGR-I regimen. The comparative results clearly show that strict adherence to the RGR-I specification has a decisive impact on task performance.

\textbf{1) RGR-I.2 Effect (Mobile-Agent-V2 vs. Mobile-Agent-E) : } Mobile-Agent-V2 and Mobile-Agent-E use the same FVP perception method to acquire environment knowledge. However, Mobile-Agent-V2 lacks task knowledge, leaving its refinement process unconstrained and reliant only on the model’s commonsense, which leads to the worst performance. Mobile-Agent-E leverages task knowledge in the form of AppTips and outperforms V2 ($CR_{UR}$ improved by 6.6\%; $CR_{KS}$ by 12.7\%). This demonstrates that the RGR-I.2 is crucial for ensuring the logical completeness of the decomposition.

\textbf{2) RGR-I.3 Effect (Mobile-Agent-E vs. Fairy) : } Both Fairy and Mobile-Agent-E rely on task knowledge. Nevertheless, Mobile-Agent-E adopts a CV-based FVP method whose perception accuracy for environment knowledge is low, causing frequent errors in the subgoal assignment stage required by RGR-I.3. In contrast, Fairy employs a UI Tree–based SSIP approach that provides high-quality, high-precision environment knowledge. This enables accurate assignment of subgoals (operations) to the correct executors (UI elements), thereby achieving the best performance ($CR_{UR}$ improved by 31.4\%; $CR_{KS}$ by 17.1\%). Fairy’s performance advantage stems precisely from its strict compliance with the RGR-I.

\begin{conclusion}{RQ2-2}
A knowledge-constrained planning engine is the cornerstone for realizing highly reliable Agentic systems. Adhering to the RGR-I ensures the logical completeness of decomposition through the use of explicit task knowledge and enables precise subgoal assignment by relying on high-quality environment knowledge, thereby significantly improving agents’ task performance and execution reliability.
\end{conclusion}

\subsubsection{\textbf{RQ2-3: Does the RGR-II Improve the User Requirement Completion Rate?}}
\begin{table}[h!]
    \centering
    \small
    \caption{Impact Of RGR-II On Agent Capabilities}
    \label{tab:key-diff-RGR-II}
    \begin{tabular}{l ccc ccc c}
        \toprule
        \multirow{2}{*}{Agent} &
        \multicolumn{3}{c}{Medium} &
        \multicolumn{3}{c}{Complex} &
        \multirow{2}{*}{Interaction Model} \\
        \cmidrule(lr){2-4} \cmidrule(lr){5-7}
        & $CR_{UR}$ & $CR_{KS}$ & $SSR$ 
        & $CR_{UR}$ & $CR_{KS}$ & $SSR$ & \\
        \midrule
        App-Agent &
        24.0\% & 43.9\% & 20.0\% &
        18.1\% & 34.0\% & 17.4\% &
        $\times$ Single-Turn \\

        Mobile-Agent-V2 &
        32.3\% & 35.2\% & 24.6\% &
        5.6\% & 31.2\% & 14.1\% &
        $\times$ Single-Turn \\

        Mobile-Agent-E &
        57.3\% & 68.4\% & 5.2\% &
        25.0\% & 62.3\% & 22.7\% &
        $\times$ Single-Turn \\

        Moba &
        39.6\% & 47.7\% & 24.6\% &
        12.5\% & 16.7\% & 18.8\% &
        $\times$ Single-Turn \\

        Fairy &
        80.2\% & 88.1\% & 15.5\% &
        70.8\% & 72.1\% & 14.7\% &
        \checkmark Multi-Turn \\

        \bottomrule
    \end{tabular}
\end{table}

To investigate RQ2-3, we evaluate the impact of the RGR-II on agents when facing ambiguous instructions from the real world. We compare Fairy against all SoTA baselines on the Medium and Complex tasks of RealMobile-Eval. All systems receive the same ambiguous user instructions provided by a Task-Driven Agent; if the agent issues a follow-up interactive query, the Task-Driven Agent will respond with clarifications based on the agent’s question and the User Requirement List. If the question cannot be answered by the User Requirement List, or if the answer could affect task outcomes, the Task-Driven Agent will refuse to answer ("I don’t know, please decide on your own.").

We select the Medium and Complex tasks in RealMobile-Eval because they are intentionally designed to include under-specified specifications and interactive ambiguity, which are necessary conditions to trigger the RGR-II. By contrast, instructions in AndroidWorld and in the Simple tasks of RealMobile-Eval are clear and thus unsuitable for evaluating this RQ. Table \ref{tab:key-diff-RGR-II} reports the full results and clearly shows that the RGR-II specification has a decisive impact on an agent’s ability to handle ambiguous tasks.

\textbf{1) Baselines (without the RGR-II)}: When facing ambiguous instructions, all baseline systems tend to perform Blind Refinement. They take liberties by guessing user intent and then successfully execute tasks with goal drift. This leads to a highly misleading outcome: their $CR_{KS}$ remains relatively stable, but $CR_{UR}$ drops sharply—by an average of 22.5\% on medium tasks and up to 42.1\% on complex tasks. This demonstrates that self-assumed execution almost inevitably deviates from the user’s true intent.

\textbf{2) Fairy (adhering to the RGR-II)}: In contrast, Fairy strictly follows the RGR-II specification, effectively mitigating the bias induced by ambiguous instructions and ensuring that the execution trajectory aligns with the user’s true intent, yielding an average $CR_{UR}$ improvement of 146.4\% over the baselines.

We further illustrate with a case: Task 28 in RealMobile-Eval involves a complex instruction. Its explicit command is: "Please use Booking to filter hotels in London with a rating of 8.5 or higher, then send the top two hotel listings to @Lemon via WeChat." The corresponding ambiguous instruction is: "Please find some good hotels in London and send a few to @Lemon." This ambiguous instruction implies several unstated user requirements: 1. "Good hotels" refers to the user's preference for selecting hotels with "rating of 8.5 or higher"; 2. "Send a few" indicates the user wants the top 2 qualifying hotels sent to @Lemon. Additionally, the instruction lacks specific application details, requiring the agent to select based on the mobile environment.

During execution, Fairy’s planning engine identifies "good hotels" (whose evaluation criteria admit multiple optional dimensions, implying an OR-decomposition) and "a few" (quantity unspecified, implying missing critical information) as under-specified Runtime Expectations. Consequently, it refuses to proceed with speculative execution, pauses execution, and activates user interaction, providing intent scaffolding (e.g., offering rating-criteria options) to request user clarification. Once feedback is obtained, these expectations are converted into clear, executable Runtime Demands, ultimately ensuring accurate fulfillment of the user’s requirements.

In contrast, agents lacking RGR-II tend to be task-success oriented in such situations and act on their own assumptions. The result almost inevitably deviates from the user’s true intent—tasks may be completed, but the actual needs of the user are not satisfied.

\begin{conclusion}{RQ2-3}
A planning engine that proactively aligns with user intent is the guarantor of highly controllable Agentic systems. By identifying Runtime Expectations at runtime and using intent scaffolding to engage the user and refine them into Runtime Demands, adherence to the RGR-II specification significantly improves the User Requirement Completion Rate under real-world ambiguous instructions.
\end{conclusion}

\subsection{RQ3: Does adhering to the EMA Improve Task Performance?}

To investigate RQ3, we evaluate the long-term performance gains brought by the EMA.
We conduct an ablation study comparing the full version of Fairy against its ablated variant, Fairy (Non-EMA). As defined in Section \ref{sec 5.3.1}, Fairy (Non-EMA) removes the long-term memory in the EMA framework (Long-term Memory is set to return empty) and skips the evolutionary execution loop of EMA-II (i.e., the self-learning process is omitted). We evaluate both on the AndroidWorld benchmark and on 10 Medium and Complex tasks in RealMobile-Eval that are highly sensitive to long-term memory. Prior to the comparative evaluation, the full Fairy executes the tasks three times to accumulate knowledge.

\begin{table}[h!]
    \centering
    \small
    \caption{The Impact Of EMA On Agent Capabilities}
    \label{tab:key-diff-EMA}
    \begin{tabular}{l cc ccc}
        \toprule
        \multirow{2}{*}{Agent} &
        \multicolumn{2}{c}{AndroidWorld} &
        \multicolumn{3}{c}{RealMobile-Eval}\\
        \cmidrule(lr){2-3} \cmidrule(lr){4-6}
        & $BSR$ & $SSR$
        & $CR_{UR}$ & $CR_{KS}$ & $SSR$\\
        \midrule
        Fairy (Non-EMA) &
        47.6\% & 20.4\% & 71.4\% & 78.3\% & 21.4\%\\

        Fairy &
        64.8\% & 14.7\% & 95.2\% & 94.2\% & 13.0\%\\

        \bottomrule
    \end{tabular}
\end{table}

Table \ref{tab:key-diff-EMA} reports the ablation results. The findings clearly indicate that the EMA is key to enabling experiential growth in agents. Because Fairy (Non-EMA) lacks the long-term memory module of  EMA-I and the evolutionary loop in EMA-II, it cannot consolidate raw experiences from Working Memory into reusable Long-term Memory. As a result, Fairy (Non-EMA) degenerates into a perpetual novice: it repeatedly makes the same mistakes in similar scenarios and continually reinvents the wheel on repetitive tasks. Moreover, it also disrupts the evolutionary sources of the task knowledge and environment knowledge on which RGR-I depends, causing task decomposition to fall back to model commonsense.
Quantitative results confirm this: on AndroidWorld, $BSR$ of Fairy (Non-EMA) drops by an average of 26.5\%, while $SSR$ rises by an average of 38.8\%. On RealMobile-Eval tasks, its $CR_{UR}$ and $CR_{KS}$ decrease on average by 25\% and 16.9\%, respectively.

The following two cases clearly illustrate the performance degradation caused by experience deficit:

\textbf{1) Case 1: Strategy optimization in an e-commerce scenario}: 

Task: "Purchase on Amazon a refined baseball cap with a rating higher than 4.0 and a price above 20 USD." With long-term memory preserved as designed under the EMA, Fairy leverages prior experience to notice the sorting/filter controls above the search results and, by clicking "25 to 30" and "Four Stars \& Up," achieves precise filtering and efficiently completes the task. After removing Long-term Memory, Fairy fails to recognize these controls and instead scrolls repeatedly to look for suitable items, markedly increasing the number of steps and the redundancy rate.

\textbf{2) Case 2: Navigation decisions in a finance application}: 

Task: "Check Alipay bills and record them." With long-term memory, Fairy quickly locates the "Me" tab based on experience and correctly navigates to the "Bills" page. Without long-term memory, Fairy mistakenly taps the Wealth Management tab and attempts to search for Bills on the wrong page, ultimately requiring many redundant steps to complete the task.

\begin{conclusion}{RQ3}
Adhering to the EMA converts each execution from a one-off consumption into an accumulable asset via a dual-loop model of execution–evolution. Through the hierarchical memory structure of EMA-I and the evolutionary loop of EMA-II, the system consolidates experience into the task knowledge and environment knowledge required by RGR-I, enabling the agent to bypass unnecessary exploration, continually optimize decisions, and significantly improve its long-term performance on repetitive and complex tasks.
\end{conclusion}

\subsection{RQ4: Does adhering to the OCA Improve system maintainability?}

To answer RQ4, we evaluate the impact of the OCA-II on system maintainability. As described in Section \ref{sec 5.3}, we conduct a human-participant study. To rigorously control inter-individual variability, we adopt a within-subjects design with counterbalanced task order. We record the average task completion time required for experts to implement a new module function similar to Fairy’s user-interaction module on App-Agent and Mobile-Agent-E, decomposed into three sub-phases.

\begin{table}[h!]
    \centering
    \small
    \caption{The Impact Of OCA On Agent Maintainability}
    \label{tab:key-diff-OCA}
    \begin{tabular}{l ccc}
        \toprule
        Maintainability &
        App-Agent & Mobile-Agent-E & Fairy
        \\
        \midrule
        Architecture &
        Single & Multi & Multi \\

        Lines Of Code &
        1.66K & 2.76K & 5.86K \\

        Files &
        12 & 8 & 71 \\

        Average Code Lines Per File &
        274 & 306 & 82 \\

        Max Code Lines In Single File &
        291 & 1036 & 261 \\

        Reading Code Time/h &
        2.9 & 8.5 & 7.1 \\

        Development Time/h &
        4.4 & 4.1 & 5.0 \\

        Debugging Time/h &
        4.8 & 12.5 & 3.2 \\
        \bottomrule
    \end{tabular}
\end{table}

Table \ref{tab:key-diff-OCA} presents the results. Because statistical power is limited by the small sample size, we treat the findings as a strong indicative trend. The results show that Fairy, which adheres to OCA-II, exhibits a maintainability advantage: its total time of development and debugging (8.2 hours) is far lower than the SoTA baselines (App-Agent: 9.2 hours; Mobile-Agent-E: 16.6 hours). This difference appears across all three sub-phases:

\textbf{1) Reading code time (assessing architectural style)}: The SoTA baselines adopt opaque coordination and communication, with highly coupled data and control flows, forcing experts to spend substantial time understanding the systems. A salient case is Mobile-Agent-E, although its architecture is simpler than Fairy’s, its tightly coupled multi-agent design makes comprehension difficult. Mobile-Agent-E uses an instantiated InfoPool class as a global shared data area (working memory) for all agents, causing extensive common coupling; moreover, its core contains multiple monolithic Python files exceeding 1,000 lines, exhibiting severe content coupling. Despite a much larger codebase, Fairy strictly follows OCA-II.2 and OCA-II.3: the Memory Bus (Memory Manager) and the Event Bus (the Citlali framework) completely separate data flow from control flow and prohibit direct inter-component calls, yielding a clear, predictable architecture and very low reading time.

\textbf{2) Development time (assessing cognitive decoupling)}: Adding new functionality (e.g., an interaction module) to SoTA baselines is invasive. Lacking OCA-II.1 cognitive decoupling, experts must modify internal logic across multiple existing components to insert new interactive behaviors, inflating development time. For App-Agent, adding a user-interaction component requires substantial changes to its prompts and context architecture, as well as new agent-orchestration logic. In Fairy, thanks to cognitive decoupling and an event-driven architecture, the expert only needs to implement a new single-responsibility interaction agent and subscribe it to relevant events (e.g., "runtime expectation detected"), without major modifications to existing components, thereby greatly reducing development time.

\textbf{3) Debugging time (assessing observability)}: The SoTA baselines are unobservable black boxes, making root-cause localization and reproduction difficult. Fairy adheres to OCA-II.2, its Memory Bus (MB) is the single source of truth for data flow (context), and its Event Bus (EB) is the sole manager of control flow. This provides white-box observability, experts can inspect MB and EB logs to effortlessly trace state changes and decision processes; once issues are found, they can rapidly build debugging scaffolds and reproduce failures by replaying events and context. This observability and debuggability effectively reduce debugging time.

\begin{conclusion}{RQ4}
Adhering to the OCA reduces functional coupling through cognitive decoupling and provides white-box observability via a state–event separation model, significantly improving maintainability, scalability, and debuggability.
\end{conclusion}

\subsection{Threats to Validity}
Our conclusions may be affected by the following potential threats. Notably, some of these are not oversights but conscious trade-offs in our experimental design. We mitigated them where possible and elaborate below.

\noindent\textbf{1) External Validity}: concerns the generalizability of our findings.

\begin{itemize}
\item \textbf{Representativeness of Benchmarks}: Our evaluation relies on AndroidWorld and our constructed RealMobile-Eval. Although RealMobile-Eval is designed to address limitations of existing benchmarks, its 30 tasks and the inclusion of specific real-world apps (e.g., Amazon, McDonald’s) may not fully represent all complex and ambiguous scenarios in mobile GUI interaction. Hence, performance of our framework on other applications or over longer task horizons remains to be further validated.

\item \textbf{Dependency on Backbone Models}: Our evaluation relies on AndroidWorld and our constructed RealMobile-Eval. Although RealMobile-Eval is designed to address limitations of existing benchmarks, its 30 tasks and the inclusion of specific real-world apps (e.g., Amazon, McDonald’s) may not fully represent all complex and ambiguous scenarios in mobile GUI interaction. Hence, performance of our framework on other applications or over longer task horizons remains to be further validated.
\end{itemize}

\noindent\textbf{2) Internal Validity}: concerns the reliability of causal inferences in our procedures.

\begin{itemize}
\item \textbf{Threat of Unifying Backbone Models}: To ensure internal validity, we unify backbones for all SoTA baselines so we compare architectures rather than model capabilities. However, this introduces a potential threat: baseline designs (e.g., prompts) may be heavily optimized for their original models, and forced substitution could harm performance due to architecture–model decoupling. A small-scale pre-study prior to the main experiments shows that performance fluctuations due to model replacement are negligible and do not alter the core ranking between SoTA baselines and Fairy. The benefits of a unified model far outweigh the potential threats it introduces.

\item \textbf{Fairness of Baseline Implementations}: Despite unifying models, existing baseline implementations (e.g., prompt engineering) may still contain unobserved confounders. We mitigate this by designing and conducting highly controlled isolation experiments (e.g., those in RQ2).

\item \textbf{Sample Size in Human-Subject Study}: In RQ4, we recruited 6 experienced experts. While prioritizing participant quality over quantity is a deliberate trade-off, the small sample size may limit statistical power, making findings (e.g., differences in code reading time) susceptible to outliers. We mitigate this by employing a more rigorous within-subjects design with counterbalanced task order to reduce individual differences.
\end{itemize}

\noindent\textbf{3) Construct Validity}: concerns whether our measurements truly capture the concept we claim to measure.
\begin{itemize}
\item \textbf{Reliability of LLM-as-Evaluator}: On RealMobile-Eval, we rely on a Task Evaluation Agent (based on GPT-4o) to compute metrics such as $CR_{UR}$. This is the largest construct-validity threat and a central trade-off of our study. We explicitly abandon traditional paradigms—manual evaluation (subjective, costly, irreproducible) and binary success rate (overly simplistic, incapable of measuring RGR-II). We argue that a rigorously validated automated evaluation pipeline (Kappa coefficient > 0.85, supplemented by human expert audits) is essential for assessing complex interactive agents: it enables large-scale, reproducible evaluation, offers analysis far beyond binary outcomes, and can genuinely evaluate ambiguous tasks closer to real-world conditions.

\item \textbf{Proxy Metrics for Maintainability}: In RQ4, we employ average task completion time and its sub-phase durations as proxy metrics for maintainability. While task duration is one of the most widely accepted proxies in empirical software engineering, it may not fully encapsulate the multifaceted construct of maintainability, which also encompasses crucial aspects such as the propensity for introducing regressions post-modification and the cognitive load imposed on developers.

\end{itemize}

\section{Related Work}
\subsection{LLM-Based Agent Reasoning Paradigms and Development Frameworks}
Many core concepts of Agentic systems originate from the classic Symbolic AI era. Wooldridge and Jennings (1995) \cite{Wooldridge1995Agents}, in their definition, summarized the core characteristics of an agent: Autonomy (acting without direct human intervention), Reactivity (perceiving and responding to the environment), Pro-activeness (initiating goal-oriented behaviors), and Social Ability (interacting with other agents). However, the Pro-activeness and Reactivity of these classic agents were severely limited by their predefined, rules-based Symbolic AI planning and knowledge capabilities. The emergence of LLMs fundamentally changed this; their powerful emergent reasoning ability gave rise to an entirely new LLM-Based reasoning paradigm.

The starting point of this new paradigm is Chain-of-Thought (CoT), proposed by Wei et al. (2022) \cite{Wei2022CoT}. As a prompting method, CoT \cite{Wei2022CoT} sparks the model's ability to solve complex reasoning tasks by guiding the LLM to generate intermediate reasoning steps, providing the possibility for the model to be transformed into an Agent with emergent planning capabilities. Subsequently, ReAct, proposed by Yao et al. (2022) \cite{Yao2023ReAct}, established the core Thought—Action—Observation loop. It combines Thought (derived from CoT \cite{Wei2022CoT}) with Action-Observation, constructing {for the first time a modern Agent paradigm capable of actively thinking, making decisions, and executing complex tasks. Although the ReAct \cite{Yao2023ReAct} loop is complete, its Thought step relies on the model's internal knowledge and lacks a mechanism for learning from failure. To this end, a series of studies have been dedicated to enhancing this loop:

\textbf{1) Knowledge Enhancement:} RAG, proposed by Lewis et al. (2020) \cite{Lewis2020RAG}, was introduced as a key technology into the agent's thought step, providing it with dynamic, verifiable declarative knowledge by connecting to external knowledge bases.

\textbf{2) Short-term Learning:} To address ReAct's \cite{Yao2023ReAct} lack of learning from failure, Reflexion, proposed by Shinn et al. (2023) \cite{Shinn2023Reflexion}, enables the agent to learn lessons from a single failure experience through a linguistic self-reflection mechanism, thereby achieving iterative self-optimization.

\textbf{3) Long-term Memory and Strategy:} To address the limitation of Reflexion \cite{Shinn2023Reflexion} being only short-term reflection, CoALA, proposed by Sumers et al. (2023) \cite{Sumers2024CoALA}, provided a cognitive blueprint that includes long-term memory. Meta-Policy Reflexion (MPR), proposed by Wu and Qu (2025) \cite{WuQu2025MPR}, took this idea a step further by extracting predicate-like universal rules, refining and solidifying fragmented experiences into meta-policies reusable across tasks.

As the underlying reasoning paradigms mature, a series of development frameworks have emerged, dedicated to integrating these capabilities to build engineering tools for practical applications. The development of these frameworks is divided into two major domains:

One category focuses on the flexible orchestration of components and multi-agent collaboration. LangChain, proposed by Chase (2022) \cite{Chase2022LangChain}, serves as a landmark in this area; it significantly lowered the development barrier through composable chain calls, tools, and memory components, rapidly becoming the de facto standard for prototyping and production applications. AutoGen, proposed by Wu et al. (2023) \cite{Wu2023AutoGen}, centers on customizable, conversational multi-agents, enabling multiple agent roles to collaboratively solve tasks through natural language negotiation.

Another category focuses on the engineering management of persistent knowledge. MemOS, proposed by Li et al. (2025) \cite{Li2025MemOS}, introduced a unified memory operating system concept, aiming to engineer the generation—organization—retrieval of long-term memory to support agents with continuous learning capabilities.

However, the work in these two domains jointly reveals a core engineering gap: on one hand, the underlying reasoning paradigms are essentially high-level behavioral guides, not rigorous engineering specifications; on the other hand, the upper-level development frameworks, as de facto implementations or specific toolboxes, are not themselves a set of universal normative methodologies.

This paper, in contrast, starts from fundamental engineering specifications, providing a systematic solution to address the aforementioned problems. OCA provides a software engineering skeleton for behavioral paradigms like ReAct \cite{Yao2023ReAct} at both macro and micro levels. RGR, based on knowledge constraints and User Interaction, ensures that the agent's behavior within the ReAct \cite{Yao2023ReAct} paradigm remains aligned with user intention based on engineering specifications. EMA provides a concrete methodology for implementing high-level cognitive blueprints like CoALA \cite{Sumers2024CoALA}, addressing the two core dimensions of how knowledge is stored and where knowledge comes from. Together, these three theories construct a complete, closed loop, from design and build to execute and interact, and finally to reflect and evolve.

\subsection{Software Engineering Architecture Paradigms}
The engineering implementation of modern Agentic systems faces severe challenges, particularly in user intention alignment, runtime self-adaptation, and complex logic layering. Although these challenges manifest in the AI domain, they are, in essence, complex software engineering problems. In the domains of software engineering and classic AI, a series of mature architectural paradigms have already provided proven theoretical references and architectural blueprints for addressing these isomorphic challenges

In addressing the problem of transforming high-level goals into low-level requirements, Goal-Oriented Requirements Engineering (GORE) \cite{vanLamsweerde2001GORE} provides a set of systematic modeling principles. Among these, the KAOS (Keep All Objectives Satisfied) framework, proposed by Dardenne et al. (1993) \cite{Dardenne1993KAOS}, offers a formal specification for the goal—sub-goal—constraint—responsibility transformation. Tropos, proposed by Bresciani et al. (2004) \cite{Bresciani2004Tropos}, advocates for integrating mental concepts (such as actor, goal, plan) throughout the entire software development lifecycle. Its extension, Tropos4AS (Tropos for Adaptive Systems) \cite{Bresciani2004Tropos}, further introduces the goal model into runtime, allowing the system to monitor goal status and guide adaptive behaviors.

In the area of system self-adaptivity, the MAPE-K loop, proposed by Kephart and Chess (2003) \cite{Kephart2003Autonomic} in the field of Autonomic Computing, is a classic reference model defined to address the self-management problems of large-scale IT systems. It defines a complete feedback loop through Monitor—Analyze—Plan—Execute, where all decisions rely on a shared Knowledge Base (K). Building on this, the Models@Runtime (M@R) paradigm, proposed by Bauer et al. (2011) \cite{Assmann2012ModelsAtRuntime}, further advocates that the system should explicitly maintain and use its own state model (i.e., the K base) at runtime, making it a first-class citizen for adaptive decision-making.

In the area of logic layering and task decomposition, Hierarchical Task Network (HTN), proposed by Erol et al. (1994) \cite{Erol1994HTN}, originating from the Symbolic AI era, is a classic planning technique. An HTN planner generates a complete execution plan by recursively decomposing high-level compound tasks into simpler sub-tasks or atomic actions (primitive actions). In distributed collaboration, the work of Wooldridge and Jennings (1995) \cite{Wooldridge1995Agents} defined the basic principles of Multi-Agent Systems (MAS) \cite{Wooldridge1995Agents}, providing a foundational theoretical framework for coordination, communication, and responsibility allocation among multiple autonomous, interacting computational entities.

Although the aforementioned paradigms have provided decades of engineering guidance for building reliable and trustworthy systems, these valuable engineering specifications are in an absent state in modern Agentic systems—which are driven by large models and centered on semantic generation and adaptive reasoning—because they have not yet been effectively adapted.

One of the core contributions of this paper is precisely to systematically reintroduce and adapt these classic SE paradigms into modern agent design: RGR borrows the specification concepts from GORE \cite{vanLamsweerde2001GORE}, innovatively transforming them from design-time to runtime to tackle the dynamic intention alignment challenges. OCA, at the macro-logic level, borrows the coordination and layering concepts from MAS \cite{Wooldridge1995Agents} and HTN \cite{Erol1994HTN} to rigorously organize multiple RGR cognitive engines and achieve micro-level observability. EMA addresses the Agent's adaptation and knowledge evolution problems by reshaping the classic MAPE-K loop \cite{Kephart2003Autonomic}, actualizing an Execution-Evolution Dual Loop. Together, these three theories systematically fuse and reshape classic SE paradigms—such as design-time specifications, hierarchical planning, and adaptive loops—into engineering specifications for modern Agentic systems. \subsection{Mobile GUI Agent} A Mobile GUI Agent is an autonomous agent that runs on a smartphone, using a LMM as its core reasoning engine to perceive GUI screens, plan tasks, and execute specific operations. The work in this domain began with AutoDroid, proposed by Wen et al. (2024) \cite{Wen2024AutoDroid}, and two mainstream research paths quickly emerged:

\textbf{1) Optimization based on Workflow and Prompt Engineering}: The first path is dedicated to achieving fine-grained control of smartphone applications by designing more advanced workflow architectures, prompt engineering, and context engineering. Closely following AutoDroid \cite{Wen2024AutoDroid}, work such as AppAgent, proposed by Li et al. (2024) \cite{Li2024AppAgentV2}, the Mobile-Agent series, proposed by Wang et al. (2025) \cite{Wang2025MobileAgentE}, MobA, proposed by Zhu et al. (2025) \cite{Zhu2025MobA}, and M3A, proposed by Rawles et al. (2024) \cite{Rawles2024AndroidWorld}, has deeply explored this direction, significantly improving the agent's task execution capability in complex applications.

\textbf{2) Optimization based on Datasets and Model Finetuning}: The second path focuses on the LMM itself, aiming to enhance the LMM's deep understanding and manipulation capability of GUI page elements and operation logic by building new, GUI-operation-specific datasets and finetuning the model. For example, CocoAgent, proposed by Ma et al. (2024) \cite{Ma2024CoCoAgent}, and MobileFlow, proposed by Nong et al. (2024) \cite{Nong2024MobileFlow}, are representative studies of this path.

The aforementioned work primarily explores Agent performance enhancement from an AI perspective (e.g., prompt optimization, workflow design, or model finetuning). However, in terms of workflow design, most of this work follows the AI domain's existing reasoning paradigms. Their architectural implementations remain ad-hoc and, when facing real-world complexity, are still prone to exposing persistent problems such as non-deterministic loss of control and insufficient observability. Fairy, built upon the software engineering framework proposed in this paper, is able to organically fuse the flexibility of modern AI reasoning paradigms with the rigor of traditional software engineering. As shown by the experimental results, Fairy achieved SoTA performance on multiple metrics. This outcome not only empirically validates the effectiveness of the framework proposed in this paper but also contributes a new high-performance baseline to the Mobile GUI Agent domain.

\section{Limitations \& Future Work}

In this paper, we proposed a systematic software engineering framework composed of Runtime Goal Refinement (RGR), Observable Cognitive Architecture (OCA), and Evolutionary Memory Architecture (EMA). We utilized this framework to construct a mobile GUI agent case study, Fairy, thereby empirically validating the framework's effectiveness in enhancing task performance, improving maintainability, and handling real-world ambiguous instructions.
Although the experimental results demonstrate the significant advantages of this framework, we must acknowledge that the framework itself possesses inherent limitations at the design level. Concurrently, limitations also exist in the definition of the framework's specifications and in our empirical validation process. This chapter will elaborate on these limitations and, based thereon, provide an outlook for future research work.

\subsection{Limitations in Framework Design}
\label{sec: limitation_in_framework_design}
The RGR, OCA, and EMA proposed in this study possess the following inherent limitations at the framework's design level:

\begin{enumerate}
    \item \textbf{RGR-I Cold Start and Knowledge Dependency:} The core of the RGR-I specification requires that the planning engine's refinement process must be explicitly constrained by task knowledge and environment knowledge. EMA, in turn, is designed to supply and iterate this knowledge via the evolution loop. This design, while logically complete, also results in the system's performance being strongly dependent on experience accumulation, i.e., a cold start problem exists. As revealed by the ablation study in RQ4, a system lacking EMA (i.e., unable to supply long-term knowledge) suffers a substantial performance degradation. This framework currently does not define a standard mechanism to address the initial knowledge acquisition problem for an agent facing entirely new tasks or environments.
    
    \item \textbf{RGR-II Collaborative Dependency on Human-in-the-Loop:} The design of the RGR-II specification prioritizes ensuring high alignment between the agent's behavior and user intention. This specification requires the system, upon identifying a runtime contingency (such as OR-decomposition or missing information), to pause autonomous execution and activate user interaction via an intent scaffold. The limitation of this design is that it strongly binds the framework's effectiveness to the availability of the human-in-the-loop. This may make the framework difficult to apply directly to scenarios requiring high autonomy, or where real-time human-computer interaction is unfeasible or inconvenient.
    
    \item \textbf{EMA Evolutionary Mechanism's A Posteriori Characteristic:}  The EMA-II specification defines the evolution loop as a post-task MAPE-K loop. This loop is responsible for reading the complete trajectory from working memory after task completion and solidifying it into long-term memory. The limitation of this design is that the agent cannot perform online learning or instantly update its long-term strategy during task execution. If the agent makes an error in the early stages of a long-cycle task, it can only rely on the execution loop for tactical correction, and cannot immediately solidify this experience into task knowledge to guide subsequent steps, which limits its immediate adaptability within long tasks.

    \item \textbf{OCA Prescriptive Overhead:}OCA provides a rigorous engineering blueprint for building white-box systems. However, this prescriptiveness comes at a cost. The OCA-I specification mandates state-control separation through cognitive decoupling and the use of an Event Bus (EB) and a Memory Bus (MB). Although the experiments in RQ2 demonstrated the maintainability advantages of this design in the long run (e.g., when adding new features), this architecture inevitably introduces high initial development costs and engineering complexity. Regarding the OCA-II specification (adopted by Fairy), its logic layering is suitable for decomposing complex tasks, but for simple tasks, this layered architecture may lead to unnecessary Over-engineering, which in turn negatively impacts system speed.
\end{enumerate}

\subsection{Limitations in Framework Specification and Empirical Validation}
\label{sec: limitations_in_the_framework_specification}
This study also possesses the following limitations at the level of specification definition and empirical scope:

\begin{enumerate}
    \item \textbf{Informal Nature of Frameworks:} The RGR, OCA, and EMA proposed in this study are, in essence, presented as a set of systematic software engineering Methodologies or Specifications, rather than as a rigorous, mathematically-based formal model. This informal characteristic, on one hand, provides necessary flexibility for engineering implementation. On the other hand, it may also lead to inconsistencies in how developers interpret and implement the specifications when applying the framework in different systems or domains, thereby affecting the framework's portability and verifiability.
    
    \item \textbf{Domain-Specific Limitation of the Empirical Study:} We selected the mobile GUI domain because its inherent challenges highly align with the core problems our framework aims to solve. The empirical results in Chapter \ref{ch:Result&Discussion} demonstrated the framework's effectiveness in this domain. However, this in-depth validation in a specific domain also constitutes a primary limitation of this study. Although we believe the engineering principles embedded within RGR, OCA, and EMA are conceptually universal, their applicability and effectiveness in other types of Agentic systems have not yet been empirically validated.
\end{enumerate}

\subsection{Future Work}
Based on the limitations in framework design, specification, and empirical validation revealed in Section \ref{sec: limitation_in_framework_design} and \ref{sec: limitations_in_the_framework_specification}, we believe future research can proceed in the following key directions to further enhance the completeness, adaptability, and universality of the RGR, OCA, and EMA.

\begin{enumerate}
    \item \textbf{Enhancing Framework Autonomy and Immediate Adaptability:} To address the inherent design limitations of RGR and EMA, future work should focus on RGR-I's cold start knowledge acquisition mechanism and EMA's online evolution loop (to complement its A Posteriori learning deficiency). This will improve the agent's immediate adaptability in long-cycle and uncertain tasks.
    
    \item \textbf{Advancing Formal Specification and Verification:} To address the problem of informal specifications, a key theoretical direction is to provide a rigorous, mathematically-based formal definition (e.g., using temporal logic or automata theory) for the RGR, OCA, and EMA. This will not only eliminate specification ambiguity but also greatly enhance the framework's verifiability and reliability, laying the theoretical foundation for building high-assurance Agentic systems.
    
    \item \textbf{Conducting Cross-Domain Empirical and Generalization Studies:} To overcome the domain-specific limitation of the empirical study, the RGR, OCA, and EMA should be extended and applied to Agentic systems beyond mobile GUI. Future empirical work should be extended to diverse domains, such as software engineering agents, embodied intelligence, or research assistant agents, to test, refine, and ultimately confirm the universality of this software engineering framework.
\end{enumerate}

\section{Conclusion}
To address the core challenges prevalent in the engineering practice of Agentic AI systems—namely, non-determinism, black-box characteristics, and the difficulty in aligning with ambiguous user intentions—this paper designs and proposes a systematic software engineering framework composed of Runtime Goal Refinement (RGR), Observable Cognitive Architecture (OCA), and Evolutionary Memory Architecture (EMA). To empirically validate the framework's effectiveness, we fully engineered it into a mobile GUI Agent named Fairy. In a series of strictly controlled experiments, Fairy demonstrated comprehensive performance advantages over current mainstream SoTA baselines: The RGR enabled it to effectively clarify ambiguous instructions through Human-in-the-loop interaction; the OCA, through logic layering, successfully decomposed complex multi-stage tasks and, through state-control separation, significantly enhanced the system's maintainability; and the EMA, via an execution-evolution dual-loop model, allowed the agent to learn from experience and continuously optimize its long-term performance. Although this framework still has limitations in formal definition and cross-domain generalization, the findings of this study confirm that the framework, as a methodology, provides an effective set of engineering specifications and a practical blueprint for building robust, observable and evolvable Agentic AI systems.



\bibliographystyle{ACM-Reference-Format}
\bibliography{reference}


\end{document}